\documentclass{article}
\pdfoutput=1
\PassOptionsToPackage{numbers, sort&compress}{natbib}
\usepackage[final]{neurips_2020}

\usepackage[utf8]{inputenc} 
\usepackage[T1]{fontenc}    
\usepackage{hyperref}       
\usepackage{url}            
\usepackage{booktabs}       
\usepackage{amsfonts,amssymb,amsthm,amsmath}
\usepackage{nicefrac}       
\usepackage{microtype}      
\usepackage{wrapfig}
\usepackage{algorithm}
\usepackage[noend]{algpseudocode}
\usepackage{xpatch}
\usepackage{textcomp}

\makeatletter
\xpatchcmd{\algorithmic}{\itemsep\z@}{\itemsep=0.5ex plus1pt}{}{}
\makeatother

\usepackage[disable,colorinlistoftodos,prependcaption,textsize=tiny]{todonotes}
\usepackage{xargs}                      
\usepackage{outlines}
\usepackage{placeins}
\usepackage{subcaption}
\usepackage{xcolor}
\usepackage{bm}
\usepackage[version=4]{mhchem}

\newcommand{\Rc}{\ensuremath \mathcal{R}}

\newcommand{\Mc}{\ensuremath \mathcal{M}}

\newcommand{\zb}{\ensuremath \mathbf{z}}
\newcommand{\mb}{\ensuremath \mathbf{m}}
\newcommand{\eb}{\ensuremath \mathbf{e}}
\newcommand{\cb}{\ensuremath \mathbf{c}}
\newcommand{\Eb}{\ensuremath \mathbf{E}}
\newcommand{\E}{\ensuremath \mathbb{E}}

\definecolor{nodeAddition}{HTML}{F19E38}%
\definecolor{buildingBlockSelection}{HTML}{6CAAC4}%
\definecolor{edgeSelection}{HTML}{47774B}%

\newcommand{\bbnode}{\textsf{\textquotesingle B\textquotesingle}}
\newcommand{\pnode}{\textsf{\textquotesingle P\textquotesingle}}
\newcommand{\interProd}{\ensuremath  \not\to_\mathcal{I}}
\newcommand{\finalProd}{\ensuremath  \not\to_\mathcal{F}}

\newcommand{\Aa}{\ensuremath{{\mathsf{\textcolor{nodeAddition}{A1}}}}}
\newcommand{\Ab}{\ensuremath{{\mathsf{\textcolor{buildingBlockSelection}{A2}}}}}
\newcommand{\Ac}{\ensuremath{{\mathsf{\textcolor{edgeSelection}{A3}}}}}

\newcommand{\MolAtL}{\ensuremath{M_{<l}}}

\newcommand{\appendixpagenumbering}{
  \break
  \pagenumbering{arabic}
  \renewcommand{\thepage}{Ap.-\arabic{page}}
}

\title{Barking up the right tree: an approach to search over molecule synthesis DAGs}

\author{%
  John Bradshaw\\
  {\small University of Cambridge}\\
  {\small MPI for Intelligent Systems}\\
  {\small \texttt{jab255@cam.ac.uk}}
  \And
  Brooks Paige\\
  {\small University College London}\\
  {\small The Alan Turing Institute}\\
  {\small \texttt{b.paige@ucl.ac.uk}}
  \And
   Matt J. Kusner\\
  {\small University College London}\\
  {\small The Alan Turing Institute}\\
  {\small \texttt{m.kusner@ucl.ac.uk}}
  \And
  Marwin H. S. Segler\\
  {\small WWU M\"unster}\\
  {\small Microsoft Research Cambridge, UK}\\
  {\small \texttt{marwin.segler@wwu.de}}
 \And
   Jos\'e Miguel Hern\'andez-Lobato\\
  {\small  University of Cambridge }\\
  {\small The Alan Turing Institute}\\
  {\small  Microsoft Research Cambridge, UK}\\
  {\small \texttt{jmh233@cam.ac.uk}}
} 

\date{}

\begin{document}
\maketitle

\begin{abstract}
  When designing new molecules with particular properties, it is not only important what to make but crucially \emph{how to make it}.
These instructions form a synthesis directed acyclic graph (DAG), describing how a large vocabulary of simple building blocks can be recursively combined through chemical reactions
to create more complicated molecules of interest. In contrast, many current deep generative models for molecules ignore synthesizability.
We therefore propose a deep generative model that better represents the real world process, by directly outputting molecule synthesis DAGs.
We argue that this provides sensible inductive biases, ensuring that our model searches over the same chemical space that chemists would also have access to, as well as interpretability.
We show that our approach is able to model chemical space well, producing a wide range of diverse molecules,
and allows for unconstrained optimization of an inherently constrained problem: maximize certain chemical properties such
that discovered molecules are synthesizable.

\end{abstract}

\section{Introduction}
\label{sect:intro}

Designing and discovering new molecules is key for addressing some of world's most pressing problems: 
creating new materials to tackle climate change, developing new medicines, and providing agrochemicals to the world's growing population.
The molecular discovery process usually proceeds in design-make-test-analyze cycles, where new molecules are designed,
made in the lab, tested in lab-based experiments to gather data, which is then analyzed to inform the next design step
\citep{holenz2016lead}.

There has been a lot of recent interest in accelerating this feedback loop, by using machine learning (ML) to design new molecules %
\citep{Gomez-Bombarelli2018-ex,segler2018generating,olivecrona2017molecular,you2018graph,Jin2018-aa,pmlr-v97-kajino19a,
assouel2018defactor,simm2019generative,simm2020reinforcement,kadurin2017cornucopia,dai2018syntax,Seff2019-dk}.
We can break up these developments into two connected goals: 
G1. \textbf{Learning strong generative models of molecules}
that can be used to sample novel molecules, for downstream screening and scoring tasks; and G2. \textbf{Molecular optimization}: Finding molecules that optimize properties of interest (e.g. binding affinity, solubility, non-toxicity, etc.).
Both these goals try to reduce the total number of design-make-test-analyze cycles needed in a complementary manner, G1 by broadening the number of molecules that can be considered in each step, and G2 by being smarter about which molecule to pick next.

Significant progress towards these goals has been made using: 
(a) representations ranging from strings and grammars \citep{segler2018generating, Gomez-Bombarelli2018-ex, Kusner2017-ry},
 to fragments \citep{Jin2018-aa, podda2020deep} and molecular graphs \citep{Simonovsky2018-md, Liu2018-ha,li2018learning}; 
and (b) model classes including latent generative models \citep{Kusner2017-ry,Gomez-Bombarelli2018-ex,De_Cao2018-sq,madhawa2019graphnvp} and
autoregressive models \citep{segler2018generating,li2018learning}.
However, in these approaches to molecular design, there is no explicit indication that the designed molecules can actually be synthesized; 
the next step in the design-make-test-analyze cycle and the prerequisite for experimental testing. 
This hampers the application of computer-designed compounds in practice, where rapid experimental feedback is essential.
While it is possible to address this with post-hoc synthesis planning
\citep{ihlenfeldt1996computer,segler2017towards,segler2018planning,szymkuc2016computer, gao2020synthesizability}, this is unfortunately very slow.

\begin{wrapfigure}{r}{0.4\textwidth}
  \centering
  \vspace{-0.5cm}
    \includegraphics[width=0.38\textwidth]{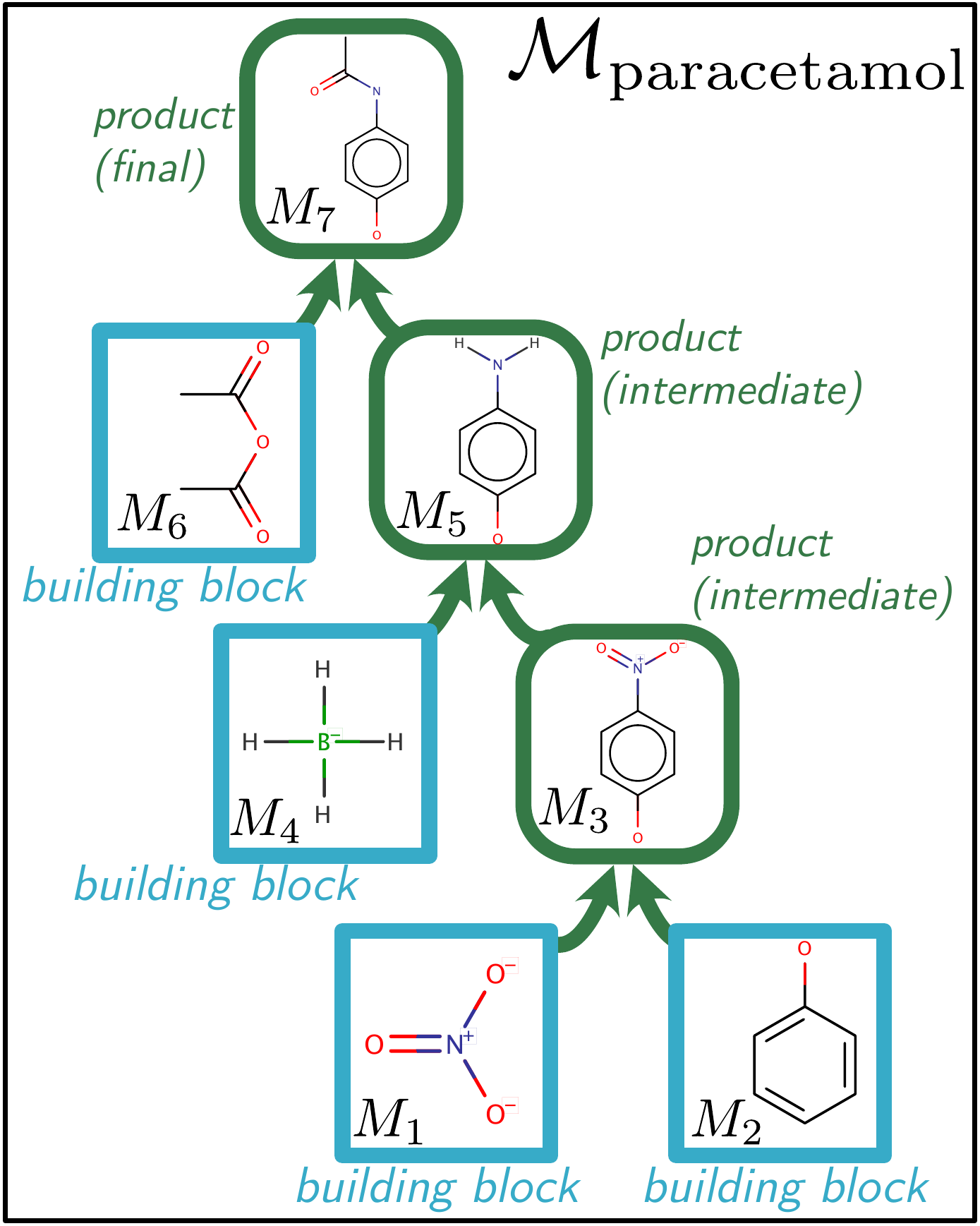}
    \caption{An example synthesis DAG for paracetamol \citep{ellis2002paracetamol}
      Note that we are ignoring some reagents, conditions and details of chirality for simplicity. 
}
    \label{fig:retrosynthesis-tree-para}
    \vspace{-0.5cm}
\end{wrapfigure}

One possibility to better link up the design and make steps, in ML driven molecular generation, is to explicitly include synthesis instructions in the design step.
Based on similar ideas to some of the discrete enumeration algorithms established in chemoinformatics, which construct molecules from building blocks via virtual chemical reactions \citep{vinkers2003synopsis,hartenfeller2012dogs,van2019virtual,chevillard2015scubidoo,humbeck2018chipmunk},
such models were recently introduced \citep{bradshaw2019model,korovina2019chembo}.
\citet{bradshaw2019model} proposed
a model to choose which reactants to combine from an initial pool, and employed an autoencoder for optimization.
However, their model was limited to single-step reactions, whereas most molecules require multi-step syntheses (e.g. see Figure \ref{fig:retrosynthesis-tree-para}).
\citet{korovina2019chembo} performed a random walk on a reaction network, deciding which molecules to assess the properties of using Bayesian optimization. However, this random walk significantly
limits the ability of that model to optimize molecules for certain properties.\footnote{Concurrent to the present work, \citet{gottipati2020learning} and \citet{horwood2020molecular} propose to optimize molecules via multi-step synthesis using reinforcement learning (RL).
However, both approaches are limited to linear synthesis trees, where intermediates are only reacted with a fixed set of starting materials,
instead of also reacting them with other intermediates. 
We believe these works, which focus more on improving the underlying RL algorithms, are complementary to the model presented here, and we envision combining them in future work.}

In this work to address this gap, we present a new architecture to generate \emph{multi-step molecular synthesis routes} and show how this model can be used for targeted optimization.
We (i) explain how one can represent synthetic routes as directed acyclic graphs (DAGs), (ii) propose a novel hierarchical neural message passing procedure that
exchanges information among multiple levels in such a synthesis DAG, and (iii) develop an efficient serialization procedure for synthesis DAGs.
Finally, we show how our approach leads to encoder and decoder networks which can both be integrated into widely used architectures and frameworks such as latent generative models (G1) \citep{kingma2013auto, rezende2014stochastic, tolstikhin2017wasserstein}, or reinforcement learning-based optimization procedures to sample and optimize (G2)  novel molecules along with their synthesis pathways. 
Compared with models not constrained to also generate synthetically tractable molecules, competitive results are obtained.

\nocite{bowman2015generating, alemi2018fixing}

\section{Formalizing Molecular Synthesis}
\label{sect:synth}

We begin by formalizing a multi-step reaction synthesis as a DAG (see Figure \ref{fig:retrosynthesis-tree-para}).%
\footnote{We represent synthesis routes as \emph{DAGs} (directed acyclic graphs) such that there is a one-to-one mapping between each molecule and each node,
even if it occurs in multiple reactions (see the Appendix for an example).} 

\paragraph{Synthesis pathways as DAGs}
At a high level, to synthesize a new molecule $M_T$, one needs to perform a series of reaction steps.
Each reaction step takes a set of molecules and physically combines them under suitable conditions to produce a new molecule.
The set of molecules that react together are selected from a pool of molecules that are available at that point.
This pool consists of a large set of initial, easy-to-obtain starting molecules (building blocks) and the intermediate products already created.
The reaction steps continue in this manner until the final molecule $M_T$ is produced, where we make the assumption
that all reactions are deterministic and produce a single primary product.

For example, consider the synthesis pathway for paracetamol, in Figure~\ref{fig:retrosynthesis-tree-para},
which we will use as a running example to illustrate our approach.
Here, the initial, easy-to-obtain building block molecules are shown in blue boxes.
Reactions are signified with arrows, which produce a product molecule (outlined in green) from reactants. 
These product molecules can then become reactants themselves.

In general, this multi-step reaction pathway forms a synthesis DAG, which we shall denote $\Mc$. 
Specifically, note that it is directed from reactants to products,
and it is not cyclic, as we do not need to consider reactions that produce already obtained reactants.

\section{Our Models}
\label{sect:model}

Here we describe a generative model of synthesis DAGs.
This model can be used flexibly as part of a larger architecture or model, as shown in the second half of this section.

\subsection{A probabilistic generative model of synthesis DAGs}

We begin by first describing a way to serialize the construction of synthesis DAGs, such that a DAG can be iteratively built up by our model predicting a series of actions.
We then describe our model, which parameterizes probability distributions over each of these actions.
Further details including specific hyperparameters along with a full algorithm are deferred to the
Appendix.\footnote{Code is provided at \url{https://github.com/john-bradshaw/synthesis-dags}.}

\begin{figure*}[t!]
  \centering
    \includegraphics[width=\textwidth]{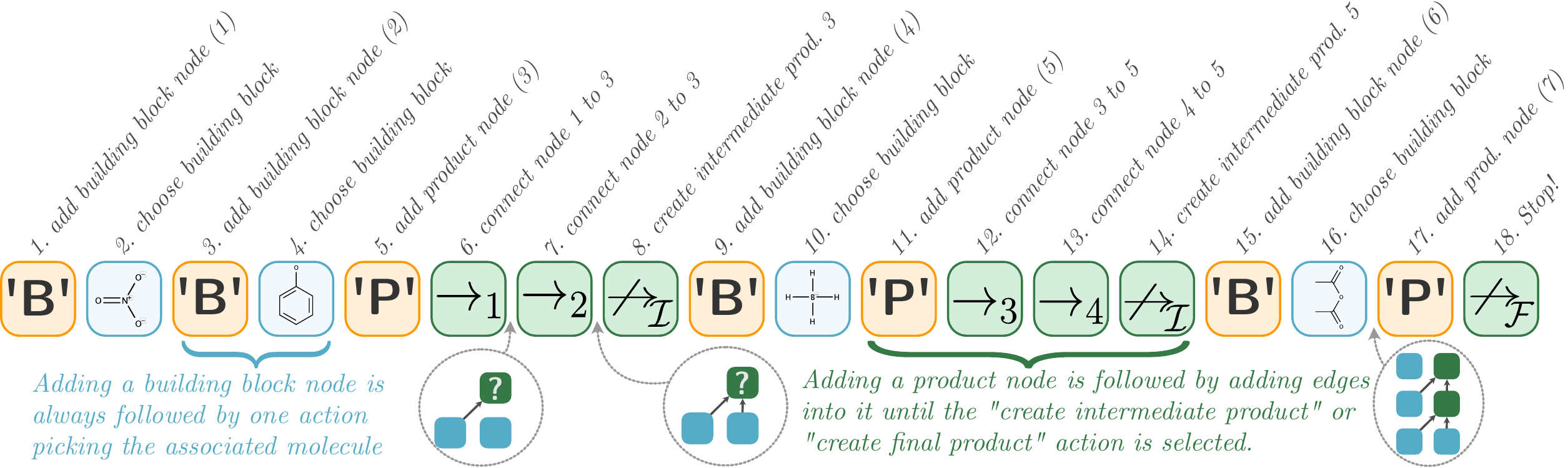}
    \caption{An example of how we can \emph{serialize} the construction of the DAG shown in Figure~\ref{fig:retrosynthesis-tree-para},
      with the corresponding DAG at that point in the sequence shown for three different time-points in the grey circles.
    The serialized construction sequence consists of a sequence of actions. These actions can be classified into belonging to three different types:
    \textcolor{nodeAddition}{(\Aa) node addition}, \textcolor{buildingBlockSelection}{(\Ab) building block molecular identity},
    and \textcolor{edgeSelection}{(\Ac) connectivity choice}. 
    By convention we start at the building block node that is furthest from the final product node, sampling randomly when two nodes are at equivalent 
    distances.
      }
    \label{fig:reconstruct-serialize}
    \vspace{-3ex}
\end{figure*}

\paragraph{Serializing the construction of DAGs}
Consider again the synthesis DAG for paracetamol shown in Figure~\ref{fig:retrosynthesis-tree-para};
how can the actions necessary for constructing this DAG be efficiently ordered into a sequence to obtain the final molecule? 
Figure~\ref{fig:reconstruct-serialize} shows such an approach.
Specifically, we divide actions into three types:
\Aa. \emph{Node-addition  (shown in yellow)}: What type of node (building block or product) should be added to the graph?; 
\Ab. \emph{Building block molecular identity  (in blue)}: Once a building block node is added, what molecule should this node represent?; 
\Ac. \emph{Connectivity choice (in green)}: What reactant nodes should be connected to a product node
(i.e., what molecules should be reacted together)? 

As shown in Figure~\ref{fig:reconstruct-serialize} the construction of the DAG then happens through a sequence of these actions.
Building block (\bbnode) or product nodes (\pnode) are selected through action type \Aa, before the identity of the molecule they 
contain is selected.
For building blocks this consists of choosing the relevant molecule through an action of type \Ab.
Product nodes' molecular identity is instead defined by the reactants that produce them, therefore action type \Ac {} is used repeatedly to either select these incoming reactant edges, or to decide to form an intermediate ($\not\to_\mathcal{I}$) or final product ($\not\to_\mathcal{F}$).
In forming a final product all the previous nodes without successors are connected up to the final product node, and the sequence is complete.

In creating a DAG, $\Mc$, we will formally denote this sequence of actions, which fully defines its structure, as $\Mc= [ V^1, V^2, V^3, \ldots, V^L ]$. 
We shall denote the associated action types as  $\bm{A} = [ A^1, A^2, A^3, \ldots, A^L ]$ with $A^l \in \{\Aa, \Ab, \Ac\}$,
and note that these are also fully defined by the previous actions chosen (e.g. after choosing a building block identity you always go back to adding a new node). 
The molecules corresponding to the nodes produced so far we will denote as $M_i$, with $i$ referencing the order in which they are created.
We will also abuse this notation and use $\MolAtL$ to describe the set of molecule nodes existing at the time of predicting action $l$.
Finally, we shall denote the set of initial, easy-to-obtain building blocks as $\Rc$.

\begin{figure*}[t!]
  \centering
    \includegraphics[width=\textwidth]{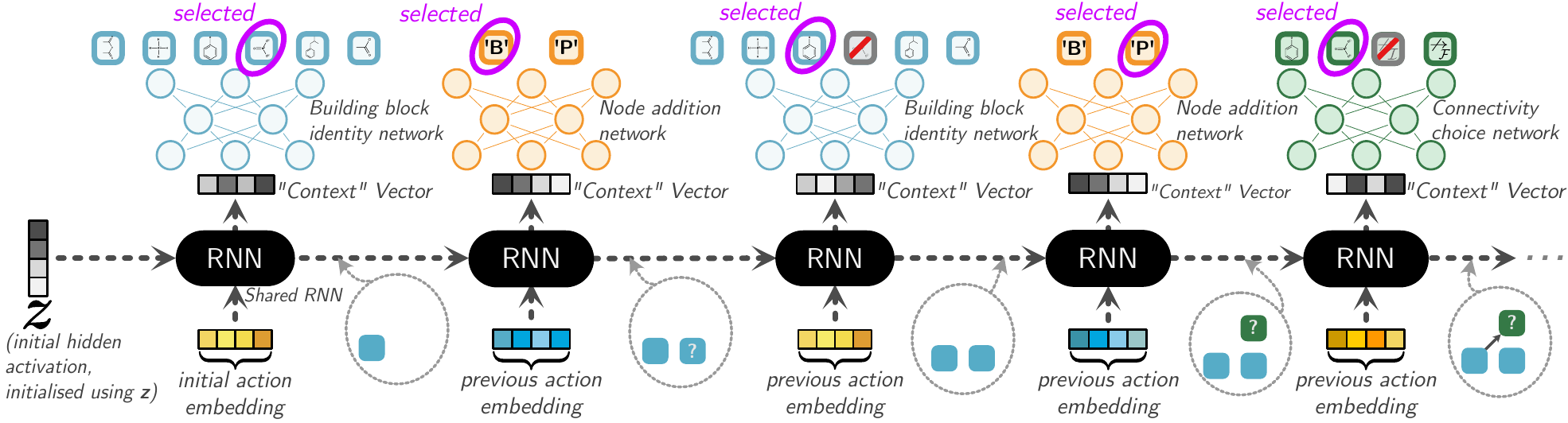}
    \caption{
      A depiction of how we use neural networks to parameterize the probability of picking actions at stages 1-6 of Figure \ref{fig:reconstruct-serialize}
      (note that as stage 1 always suggests a building block node it is automatically completed).
      A shared RNN for the different action networks receives an embedding of the previous action chosen and creates a context vector for the action network.
      When using our model as part of an autoencoder network then the initial hidden layer is parameterized by the latent space sample, $\bm{z}$.
      Each type of action network chooses a subsequent action to take (with actions that are impossible being masked out, such as selecting an already existing building block
      or creating an intermediate product before connecting up any reactants).
      The process continues until the `create final product' node is selected (see Figure~\ref{fig:reconstruct-actions-long} in the Appendix).
      Graph neural networks are used for computing embeddings of molecules where required.
    }
    \label{fig:reconstruct-actions}
\end{figure*}

\paragraph{Defining a probabilistic distribution over construction actions}
We are now ready to define a probabilistic distribution over $\Mc$. 
We allow our distribution to depend on a latent variable $\bm{z} \in \mathbb{R}^d$, the setting of which we shall ignore for now and discuss in the sections that follow.
We propose an auto-regressive factorization over the actions:
\begin{align}
  p_\theta(\Mc| \bm{z}) = \prod_{l=1}^L p_\theta(V_l | V_{<l}, \bm{z})
\end{align}
Each $p_\theta(V_l | V_{<l}, \bm{z})$ is parameterized by a neural network, with weights $\theta$. 
The structure of this network is shown in Figure~\ref{fig:reconstruct-actions}. 
It consists of a shared RNN module that computes a `context' vector. This `context' vector then gets fed into a feed forward action-network for predicting each action.
A specific action-network is used for each action type, and the action type also constrains the actions that can be chosen from this network:
$V^l_{|A^l=\Aa} \in \{\bbnode, \pnode \}$, $V^l_{|A^l=\Ab} \in \Rc$, and $V^l_{|A^l=\Ac} \in \MolAtL \cup \{\not\to_\mathcal{I}, \not\to_\mathcal{F}\}$. 
The initial hidden layer of the RNN is initialized by $\bm{z}$ and we feed in the previous chosen action embedding as input to the RNN at each step.
Please see the Appendix for further details including pseudocode for the entire generative procedure in detail as a probabilistic program  (Alg.~\ref{alg:generator}).

\paragraph{Action embeddings}
For representing actions to our neural network we need continuous embeddings, $\bm{h} \in \mathbb{R}^d$.
Actions can be grouped into either (i)  those that \emph{select a molecular graph} (be that a building block, $g \in \Rc$, or a
reactant already created, $g' \in \MolAtL$); or (ii) those that perform
a more \emph{abstract action} on the DAG, such as creating a new node ($\bbnode$, $\pnode$),
producing an intermediate product ($\not\to_\mathcal{I}$), or lastly producing a final product ($\not\to_\mathcal{F}$).
For the abstract actions the embeddings we use are each distinct learned vectors that are parameters of our model.

With actions involving a molecular graph, instead of learning embeddings for each molecule we compute embeddings that take the structure of the molecular graph into account.
Perhaps the simplest way to do this would be using fixed molecule representations such as Morgan fingerprints \citep{Rogers2010-tv,morgan1965generation}.
However, Morgan fingerprints, being fixed, cannot learn which characteristics of a molecule are important for our task.
Therefore,  for computing molecular embeddings we instead choose to use deep graph neural networks (GNN), specifically Gated Graph Neural Networks \citep{li2015gated} (GGNNs).
GNNs can learn which characteristics of a graph are important and have been shown to perform well on a variety of tasks involving small organic molecules \citep{gilmer2017neural,duvenaud2015convolutional}.

\paragraph{Reaction prediction}
When forming an intermediate or final product we need to obtain the molecule that is formed from the reactants chosen.
At training time we can simply fill in the correct molecule using our training data.
However at test time this information is not available.
We assume however we have access to an oracle, $\mathsf{Product}(\cdot)$, that given a set of reactants produces the major product formed
(or randomly selects a reactant if the reaction does not work).
For this oracle we could use any reaction predictor method \citep{kayala2011learning,kayala2012reactionpredictor,wei2016neural,coley2017prediction,
Jin2017-hh,fooshee2018deep,bradshaw2019generative,coley2019graph,do2019graph,segler2017neural,schwaller2018molecular}.
In our experiments we use the Molecular Transformer \citep{schwaller2018molecular}, as it currently obtains state-of-the-art performance
on reaction prediction \citep[Table 4]{schwaller2018molecular}. Further details of how we use this model are in the Appendix.

It is worth pointing out that current reaction predictor models are not perfect, sometimes predicting incorrect reactions.
By treating this oracle as a black-box however, our model can take advantage of any future developments of these methods or use alternative 
predictors for other reasons, such as to control the reaction classes used.

\subsection{Variants of our model}

Having introduced our general model for generating synthesis DAGs of (molecular) graphs (DoGs), we detail two
variants: an autoencoder (DoG-AE) for learning continuous
embeddings (G1), and a more basic generator (DoG-Gen) for performing molecular optimization via fine-tuning or reinforcement learning (G2).

\subsubsection{DoG-AE: Learning a latent space over synthesis DAGs} 
For G1, we are interested in learning mappings between  molecular space and latent continuous embeddings.
This latent space can then be used for exploring molecular space via interpolating between two points,
as well as sampling novel molecules for downstream screening tasks.
To do this we will use our generative model as the \emph{decoder} in an autoencoder structure, which we shall call DoG-AE. 
We specify a Gaussian prior over our latent variable $\bm{z}$, where each $\bm{z} \sim p(\bm{z})$ can be thought of as describing 
different types of synthesis pathways.

The autoencoder consists of a stochastic mapping from $\zb$ to synthesis DAGs and a separate \emph{encoder}, $q_\phi(\zb \mid \Mc)$
a stochastic mapping from synthesis DAGs, $\Mc$, to latent space.
Using our generative model of synthesis DAGs, described in the previous subsection, as our decoder
(with the latent variable initializing the hidden state of the RNN as shown in Figure~\ref{fig:reconstruct-actions}),  we are left to define both our autoencoder loss
and encoder.

\paragraph{Autoencoder loss}
We choose to optimize our autoencoder model using a Wasserstein autoencoder (WAE)
objective with a negative log likelihood cost function \cite{tolstikhin2017wasserstein},
\begin{align}
  \min_{\phi,\theta}&\; \E_{\Mc \sim p(\Mc)} \E_{q_\phi(\zb|\Mc)} \Big[ - \log p_\theta(\Mc \mid \zb) \Big] + \lambda \mathcal{D}(q_\phi(\zb), p(\zb)),\label{eq:autoencoder}
\end{align}
where following  \citet{tolstikhin2017wasserstein} 
$\mathcal{D}(\cdot, \cdot)$ is a maximum mean discrepancy (MMD) divergence measure \citep{gretton2012kernel} and $\lambda=10$;
alternative autoencoder objectives could easily be used instead.

\begin{figure*}[t]
  \centering
    \includegraphics[width=\textwidth]{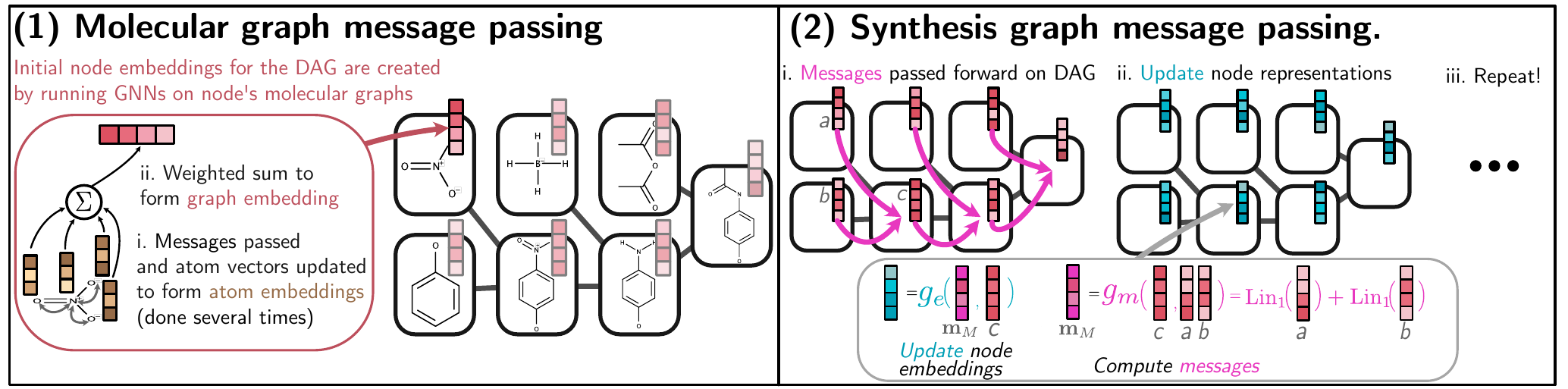}
    \caption{The encoder embeds the DAG of Graphs (DoG) into a continuous latent space.
    It does this using a two-step hierarchical message passing procedure.
    In step 1 (Molecular graph message passing) it computes initial embeddings for the DAG nodes
    by forming graph-level embeddings using a GNN on the molecular graph associated with each node.
    In step 2 (Synthesis graph message passing) a message-passing algorithm is again used, however, this time on the synthesis DAG itself, passing
    messages forward.
    In our experiments we use GGNNs \citep{li2015gated} for both message passing steps (see the Appendix for further details).
    The final representation of the DAG is taken from the node embedding of the final product node.
    }
    \label{fig:encoder}
    \vspace{-0.5cm}
\end{figure*}

\paragraph{Encoder} At a high level, our encoder consists of a two-step hierarchical message passing procedure described in Figure~\ref{fig:encoder}:
(1) Molecular graph message passing; (2) Synthesis graph message passing. 
Given a synthesis DAG $\Mc$, each molecule $M$ within that pathway can itself be represented as a graph:
each atom $a$ is a node and each bond $b_{a,a'}$ between atoms $a$ and $a'$ is an edge. 
As we did for the decoder, we can embed these molecular graphs into continuous space using any graph neural network, 
in practice we choose to use the same network as the decoder (also sharing the same weights).

Given these initial molecular graph embeddings $\eb_M^0$ (now node embeddings for the DAG), 
we would like to use them to embed the entire synthesis DAG $\Mc$.
We do so by passing the molecular embeddings through another second step of message passing (the synthesis graph message passing), 
this time across the synthesis DAG. Specifically, we use the message function $g_m$ to compute messages 
as $\mb_M^t \!=\! g_m(\eb_M^{t-1}, \eb_{N(M)}^{t-1})$, where $N(M)$ are the predecessor molecules
in $\Mc$ (i.e., the reactants used to create $M$), before updating the DAG node embeddings with this message using the
function $\eb_M^t \!=\! g_e(\mb_M^t, \eb_M^{t-1})$.
After $T'$ iterations we have the final DAG node embeddings $\eb_M^{T'}$,
which we aggregate as $\boldsymbol{\mu} \!=\! g_\mu(\Eb_{\Mc}^{T'})$ and $\boldsymbol{\sigma}^2 \!=\! \exp(g_{\log \sigma^2}(\Eb_{\Mc}^{T'}))$.
These parameterize the mean and variance of the encoder
as: $q_\phi(\zb \mid \Mc) \!:=\! \mathcal{N}(\boldsymbol{\mu}, \boldsymbol{\sigma}^2)$.

We are again flexible about the architecture details of the message passing procedure
(i.e., the architecture of the GNN functions $g_m,g_e,g_\mu, g_{\log \sigma^2}$).
We leave these specifics up to the practitioner and describe in the Appendix
details about the exact model we use in our experiments, where similar to the GNN used for the molecular embeddings,
we use a GGNN \citep{li2015gated} like architecture.

\subsubsection{DoG-Gen: Molecular optimization via fine-tuning}
For molecular optimization, we consider a model trained without a latent space;
we use our probabilistic generator of synthesis DAGs and fix $\zb\!=\!\boldsymbol{0}$, we call this model DoG-Gen.
We then adopt the hill-climbing algorithm from \citet[\S 7.4.6]{brown2019guacamol} \citep{segler2018generating,neil_exploring_2018}, 
an example of the cross-entropy method \citep{De_Boer2005-nc}, or also able to be viewed as a variant of
REINFORCE \citep{Williams1992-ov} with a particular reward shaping (other RL algorithms could also be considered here).
For this, our model is first  pre-trained via maximum likelihood to match the training dataset distribution $p(\Mc)$.
For optimization, we can then fine-tune the weights
$\theta$ of the decoder: this is done by sampling a large number of
candidate DAGs from the model, ranking them according to
a target, and then fine-tuning our model’s weights on the top K
samples (see Alg.~\ref{alg:fine-tuning} in the Appendix for full pseudocode of this procedure).

\section{Experiments}
\label{sect:experiments}

We now evaluate our approach to generating synthesis DAGs on two connected goals set out in the introduction:
(G1) can we model the space of synthesis DAGs well and using DoG-AE interpolate within that space, and 
(G2) can we find optimized molecules for particular properties using DoG-Gen and the fine-tuning technique set out in the previous section.
To train our models, we create a dataset of synthesis DAGs based on the USPTO reaction dataset
\citep{lowe2012extraction}. We detail the construction of this dataset in the Appendix (\S\ref{sect:DatasetCreation}).
We train both our models on the same dataset and find that DoG-AE obtains a reconstruction accuracy (on our held out test set) of 65\% when greedily decoding
(i.e. picking the most probable action at each stage of decoding).

\subsection{Generative modeling of synthesis DAGs}

We begin by assessing properties of the \emph{final} molecules produced by our generative model of synthesis DAGs. 
Ignoring the synthesis allows us to compare against previous generative models for molecules including 
SMILES-LSTM (a character-based autoregressive language model for SMILES strings)\citep{segler2018generating},
the Character VAE (CVAE) \citep{Gomez-Bombarelli2018-ex}, the Grammar VAE (GVAE) \citep{Kusner2017-ry},
the GraphVAE \citep{Simonovsky2018-md}, the Junction Tree Autoencoder (JT-VAE) \citep{Jin2018-aa},
the Constrained Graph Autoencoder (CGVAE) \citep{Liu2018-ha},
and Molecule Chef \citep{bradshaw2019model}.%
\footnote{We reimplemented the CVAE and GVAE models in PyTorch and found that our implementation is significantly better than \citep{Kusner2017-ry}'s published results.
We believe this is down to being able to take advantage of some of the latest techniques for training these models
(for example $\beta$-annealing\citep{higgins2017beta, alemi2018fixing}) as well as hyperparameter tuning.} 
Further details about the baselines are in the Appendix.
These models cover a wide range of approaches for modeling structured molecular graphs.
Aside from Molecule Chef, which is limited to one step reactions and so is unable to cover as much of molecular space as our approach, these other 
baselines do not provide synthetic routes with the output molecule.

As metrics we report those proposed in previous works \citep{segler2018generating,Liu2018-ha,Kusner2017-ry,bradshaw2019model,Jin2018-aa}.
Specifically, validity measures how many 
of the generated molecules can be parsed by the chemoinformatics software RDKit \citep{rdkit}.
Conditioned on validity, we consider the proportions of generated molecules that are
unique (within the sample), novel (different to those in the training set), and pass the quality filters (normalized relative to the training set)
proposed in \citet[\S3.3]{brown2019guacamol}. Finally we measure the Fréchet ChemNet Distance (FCD) \citep{doi:10.1021/acs.jcim.8b00234} between
generated and training set molecules. 
We view these metrics as useful sanity checks, showing that sensible molecules are produced, albeit with limitations
\citep{brown2019guacamol}, \cite[\S 2]{renz2020failure}. We include additional checks in the Appendix.

Table~\ref{table:generation-properties} shows the results.
Generally we see that many of these models perform comparably with no model performing better than all of the others on all of the tasks.
The baselines JT-VAE and Molecule Chef have relatively high performance across all the tasks, although by looking at the FCD score
it seems that the molecules that they produce are not as close to the original training set as those suggested by the simpler character 
based SMILES models, CVAE or SMILES LSTM.
Encouragingly, we see that the models we propose provide comparable scores to many of the models that do not provide synthetic routes and achieve
good performance on these sanity checks.

\begin{table}[t]
  \begin{center}
    \caption{
Table showing the percentage of valid molecules generated and then conditioned on this the
uniqueness, novelty and normalized quality \cite[\S3.3]{brown2019guacamol} (all as \%, higher better)
as well as FCD score (Fréchet ChemNet Distance, lower better) \citep{doi:10.1021/acs.jcim.8b00234}.
For each model we generate the molecules by decoding from 20k prior samples from the latent space.
}

\begin{tabular}{lrrrrr}
    \toprule  
    Model Name    & \multicolumn{1}{l}{Validity ($\uparrow$)} & \multicolumn{1}{l}{Uniqueness ($\uparrow$)} & \multicolumn{1}{l}{Novelty ($\uparrow$)} &
    \multicolumn{1}{l}{Quality ($\uparrow$)} & \multicolumn{1}{l}{FCD ($\downarrow$)} \\
  \midrule
DoG-AE        & 100.0                       & 98.3                           & 92.9                       & 95.5                       & 0.83                    \\
DoG-Gen       & 100.0                       & 97.7                          & 88.4                       & 101.6                      & 0.45                    \\
\midrule
Training Data & 100.0                       & 100.0                         & 0.0                        & 100.0                      & 0.21 \\ 
\midrule
SMILES LSTM \citep{segler2018generating}  & 94.8                        & 95.5                          & 74.9                       & 101.93                      & 0.46                   \\
CVAE     \citep{Gomez-Bombarelli2018-ex}     & 96.2                        & 97.6                          & 76.9                       & 103.82                      & 0.43                    \\
GVAE     \citep{Kusner2017-ry}     & 74.4                        & 97.8                           & 82.7                       & 98.98                       & 0.89                    \\
GraphVAE  \citep{Simonovsky2018-md}     & 42.2                        & 57.7                          & 96.1                       & 94.64                       & 13.92                   \\
JT-VAE   \citep{Jin2018-aa}     & 100.0                       & 99.2                          & 94.9                       & 102.34                      & 0.93                    \\
CGVAE      \citep{Liu2018-ha}    & 100.0                       & 97.8                          & 97.9                       & 45.64                       & 14.26                   \\
Molecule Chef \citep{bradshaw2019model} & 98.9                        & 96.7                          & 90.0                       & 99.0                       & 0.79               \\
   \bottomrule
\end{tabular}
\label{table:generation-properties}

\end{center}
\end{table} 

\begin{figure}[t]
\centering
\includegraphics[width=0.9\textwidth]{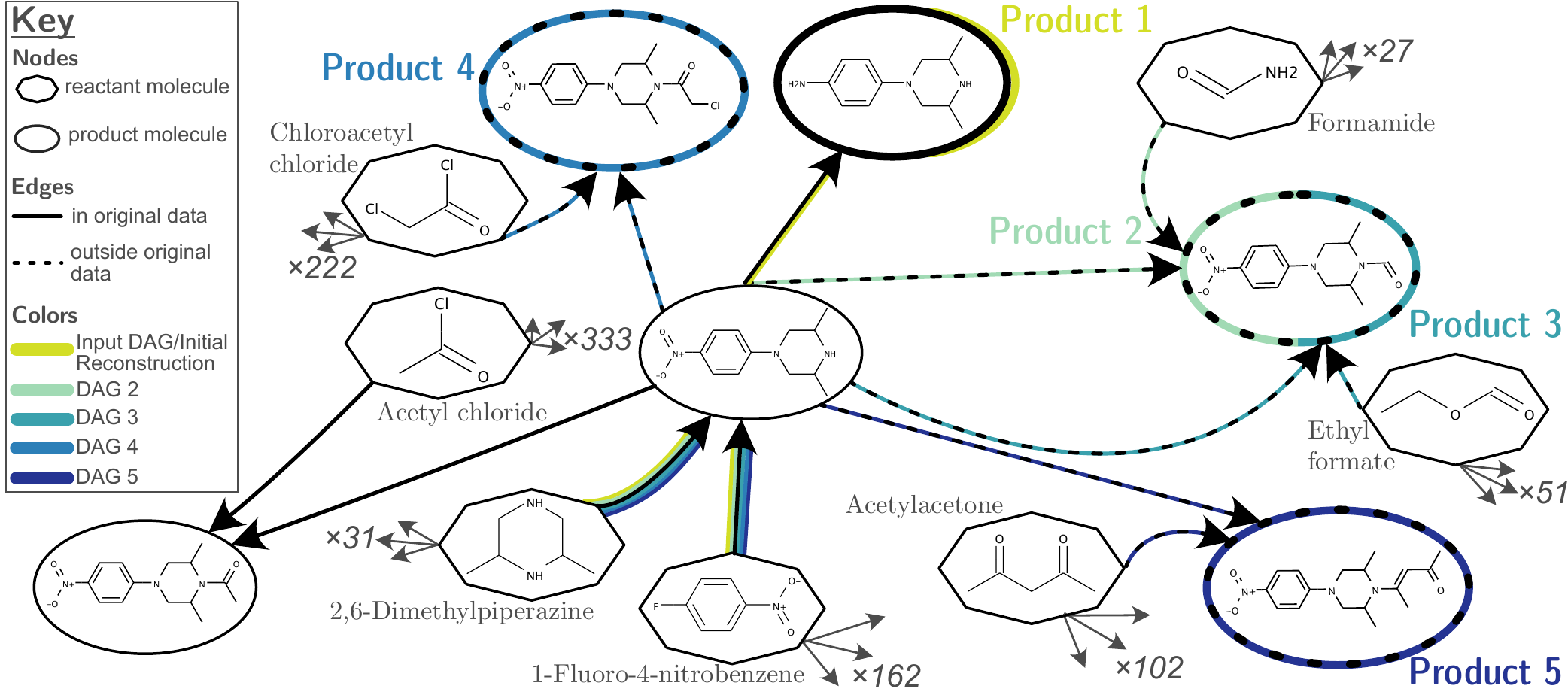}
\caption{
  Using a variant of the DoG-AE model, as we randomly walk in the latent space we decode out to similar DAGs nearby, 
  unseen in training. 
  Reactions and nodes that exist in our original dataset are outlined in solid lines, whereas those that have been discovered
by our model are shown with dashed lines.
}\vspace{-0.5cm}
\label{fig:randomwalk1}
\end{figure}

\paragraph{The latent space of synthesis DAGs.}
The advantage of our model over the others is that it directly generates synthesis DAGs, indicating how a generated molecule could be made.
To visualize the latent space of DAGs we start from a training 
synthesis DAG and walk randomly in latent space until we have output five different synthesis DAGs. We plot the combination of these DAGs, which can be seen as a reaction network, in Figure \ref{fig:randomwalk1}. 
We see that as we move around the latent space many of the synthesis DAGs have subgraphs that are isomorphic, resulting in similar final molecules.

\subsection{Optimizing synthesizable molecules}

We next look at how our model can be used for the optimization of molecules with desirable properties.
To evaluate our model, we compare its performance on a series of 10 optimization tasks from GuacaMol \citep[\S 3.2]{brown2019guacamol} 
against the three best reported models \citet[Table 2]{brown2019guacamol} found:
(1) SMILES LSTM \citep{segler2018generating}, which does optimization via fine-tuning;
(2) GraphGA \citep{jensen2019graph}, a graph genetic algorithm (GA);
and (3) SMILES GA \citep{yoshikawa2018population}, a SMILES based GA.
We train all methods on the same data, which is derived from USPTO and, as such, should give a strong bias for synthesizability.

We note that we should not expect our model to find the best molecule if judged solely on the GuacaMol task score;
our model has to build up molecules from set building blocks and pre-learned reactions,
which although reflecting the real-life process of molecule design, means that it is operating in a more constrained regime.
However, the final property score is not the sole factor that is important when considering a proposed molecule. 
Molecules also need to: (i) exist without degrading or reacting further (\emph{i.e., be sufficiently stable}),
and (ii) be able to actually be created in practice (\emph{i.e., be synthesizable}).
To quantify (i) we consider using the quality filters proposed in \citet[\S 3.3]{brown2019guacamol}.
To quantify (ii) we use Computer-Aided Synthesis Planning \citep{boda2007structure,segler2018planning,gao2020synthesizability}.  
Specifically, we run a retrosynthesis tool on each molecule to see if a synthetic route can be found, and if so how many steps are involved
\footnote{Note, like reaction predictors, these tools are imperfect, but still (we believe) offer a sensible current method for evaluating our models.}.
We also measure an aggregated synthesizability score over each step (see \S\ref{sect:appSynScore} in the Appendix), with a higher synthesizability score indicating that the individual
reactions are closer to actual reactions and so hopefully more likely to work.  
All results are calculated on the top 100 molecules found by each method for each GuacaMol task.

\begin{figure}[t]
\centering
\includegraphics[width=\textwidth]{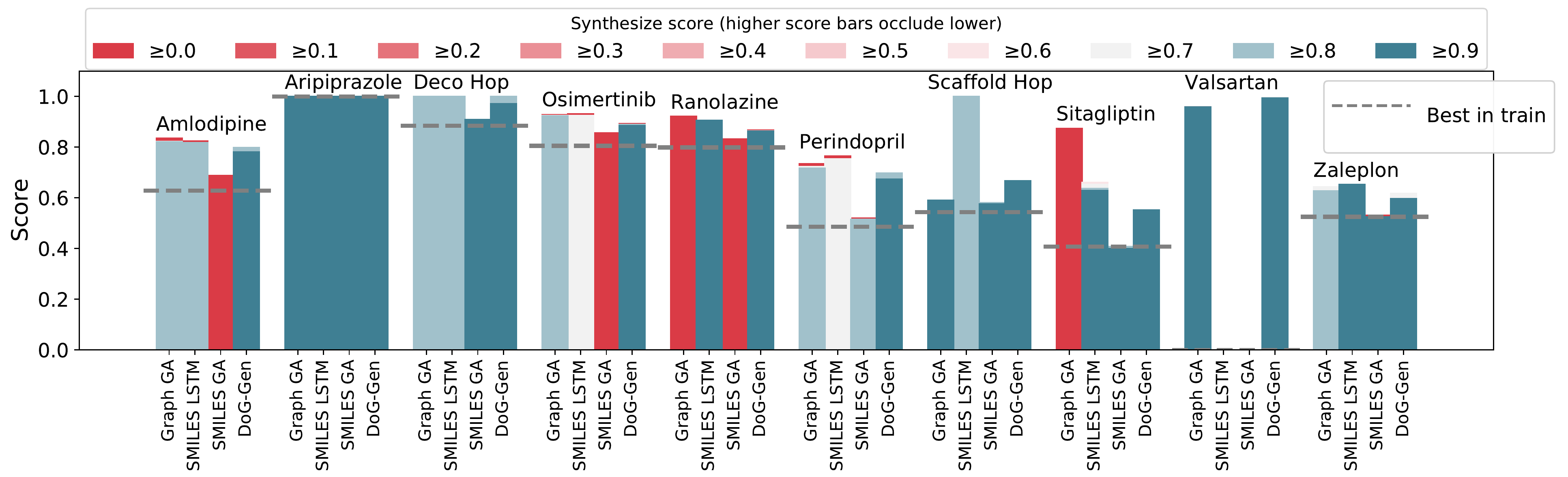}
\caption{
The score of the best molecule found by the different approaches over a series of ten GuacaMol benchmark tasks \citep[\S 3.2]{brown2019guacamol},
with the task name labeled above each set of bars.
GuacaMol molecule scores (y-axis) range between 0 and 1, with 1 being the best. 
We also use colors to indicate the synthesizability score of the best molecule found.
Note that bars representing a molecule within a higher synthesizability score bucket (e.g blue)
will occlude lower synthesizability score bars (e.g. red).
The dotted gray lines represent the scores of the best molecule in our training set.
}
\label{fig:benchmark_scores_all_thresh}
\vspace{-1em}
\end{figure}

The results of the experiments are shown in Figures \ref{fig:benchmark_scores_all_thresh} and \ref{fig:quality_filters} 
and Table~\ref{table:aggr_synth_stats} (see the Appendix for further results). 
In Figure \ref{fig:benchmark_scores_all_thresh} we see that, disregarding synthesis, 
in general Graph GA and SMILES LSTM produce the best scoring molecules for the GuacaMol tasks.  
However, corroborating the findings of \citep[FigureS6]{gao2020synthesizability}, we note that the GA methods regularly suggest molecules where no synthesis routes can be found. 
Our model, because it decodes synthesis DAGs, consistently finds high-scoring molecules while maintaining high synthesis scores.
Furthermore, Figure \ref{fig:quality_filters} shows that a high fraction of molecules produced by our method pass the quality checks (consistent with the 
amount in the training data), whereas for some of the other methods the majority of molecules can fail.

\begin{figure}[t]
\centering
\includegraphics[width=\textwidth]{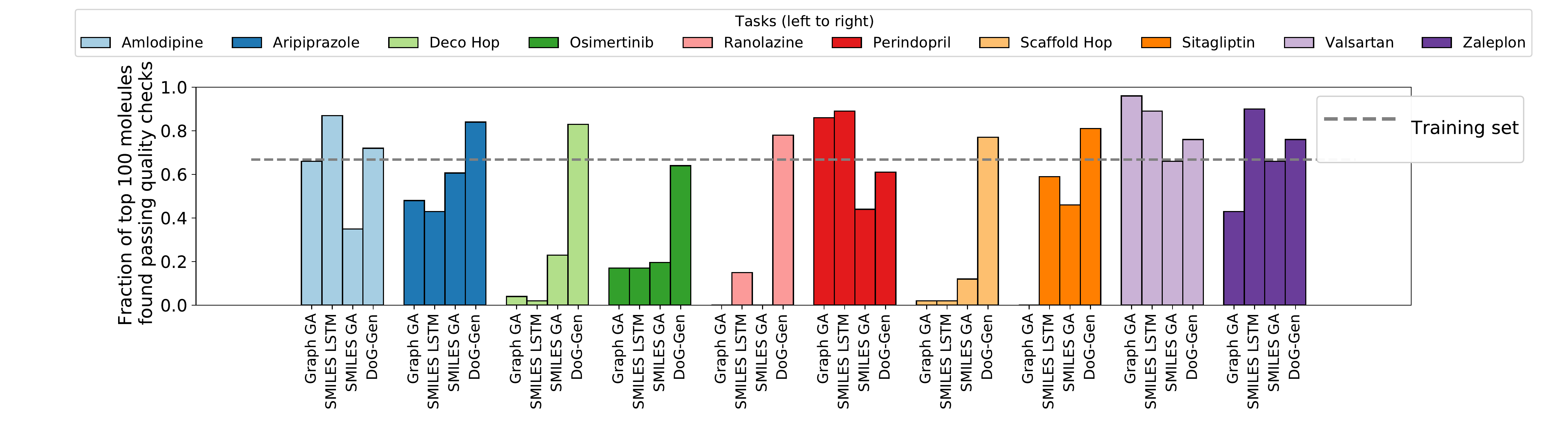}
\caption{
  The fraction of the top 100 molecules proposed that pass the quality filters, over a series of ten GuacaMol tasks \citep[\S 3.2]{brown2019guacamol}.
  The fraction of molecules in our initial training set (used for training the models) that pass the filters is shown by the dotted gray line.
}
\label{fig:quality_filters}
\end{figure}

\begin{table}
\centering
\caption{
  Table showing metrics quantifying the stability and synthesizability of the molecules suggested by each method for the GuacaMol optimization tasks.
The metrics we use include: the fraction of molecules for which a synthetic route is found, the mean synthesizability score,
the median number of synthesis steps (for synthesizable molecules), and the fraction of molecules that pass the quality filters from
\citet[\S 3.3]{brown2019guacamol}.
The metrics are computed using the aggregation of the top 100 molecules for each GuacaMol task over all of the ten GuacaMol tasks we consider.
}
    \label{table:aggr_synth_stats}

\begin{tabular}{lrrrr}
\toprule
             &   Frac. Synthesizable ($\uparrow$) &   Synth. Score ($\uparrow$)  &
   Median \# Steps ($\downarrow$) & Quality ($\uparrow$)   \\ \midrule
  DoG-Gen     &                  0.9  &                0.76 &                4  & 0.75 \\
  \midrule
  Graph GA    &                  0.42 &                0.33 &                6 & 0.36  \\ 
  SMILES LSTM &                  0.48 &                0.39 &                5 & 0.49  \\ 
  SMILES GA   &                  0.29 &                0.25 &                3 & 0.39  \\ 
  \bottomrule
\end{tabular}
\end{table}

\section{Related Work}
\label{sect:related}


In computer-aided molecular discovery there are two main approaches for choosing the next molecules to evaluate, 
virtual screening \citep{walters1998virtual,shoichet2004virtual,pyzer2015high,van2019virtual,walters2018virtual,reymond2012enumeration,chevillard2015scubidoo}
and de novo design \citep{hartenfeller2011enabling, schneider2013novo}.
De novo design, whilst enabling the searching over large chemical spaces, can lead to unstable and unsynthesizable molecules \citep{hartenfeller2011enabling, brown2019guacamol,gao2020synthesizability}.
Works to address this in the chemoinformatics literature have built algorithms around virtual reaction schemes \citep{vinkers2003synopsis,hartenfeller2012dogs}.
However, with these approaches there is still the problem of how best to search through this action space,
a difficult discrete optimization problem, and so these algorithms often resorted to optimizing greedily one-step at a time or using GAs.

Neural generative approaches \citep{Gomez-Bombarelli2018-ex, segler2018generating, olivecrona2017molecular}  have attempted to tackle this latter problem by
designing models to be used with  Bayesian optimization or RL based search techniques. However, these approaches have often lead to molecules containing
unrealistic moieties \citep{brown2019guacamol, gao2020synthesizability}. Extensions have
investigated representing molecules with grammars \citep{Kusner2017-ry, dai2018syntax}, or explicitly constrained graph representations
\citep{Jin2018-jf,Jin2019-gm,Jin2018-aa,samanta2019nevae,Liu2018-ha, podda2020deep}, which while fixing the realism somewhat, still ignores synthesizability.

ML models for generating or editing graphs (including DAGs) have also been developed for other application areas, such as citation networks, community graphs, network 
architectures or source code \citep{zhang2019d, chen2018tree, alvarez2016tree, chakraborty2018codit, pmlr-v80-you18a}. 
DAGs have been generated in both a bottom-up (e.g. \citep{zhang2019d}) and top-down  (e.g. \citep{chen2018tree}) manner.
Our approach is perhaps most related to \citet{zhang2019d},
which develops an autoencoder model for DAGs representing neural network architectures.
However, synthesis DAGs have some key differences to those typically representing neural network architectures or source code,
for instance nodes represent molecular graphs and should be unique within the DAG.

\section{Conclusions}

In this work, we introduced a novel neural architecture component for molecule design, which
by directly generating synthesis DAGs alongside molecules, captures how molecules are made in the lab. 
We showcase how the component can be mixed and matched in different paradigms, 
such as WAEs and RL, demonstrating competitive performance on various benchmarks.

\newpage

\section*{Broader Impact}
Molecular de novo design, the ability to faster discover new advanced materials,
could be an important tool in addressing many present societal challenges, such as global health and climate change. 
For example, it could contribute towards a successful transition to clean energy,
through the development of new materials for energy production (e.g. organic photovoltaics) and storage (e.g. flow batteries).
We hope that our methods, by producing synthesizable molecules upfront, are a contribution to the research in this direction.

An application area for molecule de novo design we are particularly enthusiastic about and believe could lead to large positive societal outcomes is drug design.
By augmenting the capabilities of researchers in this area we can reduce  the cost of discovering new drugs for treating diseases.
This may be particularly helpful in the development of treatments for neglected tropical diseases or orphan diseases, in which currently there are
often poorer economic incentives for developing drugs.

Whilst we are excited by these positive benefits that faster molecule design could bring, it is also important to be mindful of possible risks.
We can group these risks into two categories, (i) use of the technology for negative \emph{downstream applications}
(for instance if our model was used in the design of new chemical weapons), and 
(ii) negative \emph{side effects} of the technology (for instance including the general downsides of increased automation, such as an increased chance of accidents).
If not done carefully, increased automation can have a negative impact on jobs, and in our particular case may lead to a reduction in the control
and understanding of the molecular design process. 
We hope that this can be mitigated by communicating clearly about near-term capabilities, using and further developing sensible benchmarks/metrics,
as well as the development of tools to better explain the ML decision making process.

\begin{ack}
This work was supported by The Alan Turing Institute under the EPSRC grant EP/N510129/1. 
We are also grateful to The Alan Turing Institute for providing compute resources.  
JB also acknowledges support from an EPSRC studentship.
We are thankful to Gregor Simm and Bernhard Sch\"olkopf for helpful discussions.
\end{ack}

\FloatBarrier

\bibliographystyle{plainnat}
\bibliography{refs}

\FloatBarrier
\newpage

\appendixpagenumbering
\appendix
\section*{\huge Appendix}


This Appendix is broken up into several sections:

\begin{description}
  \item[Section~\ref{sect:synDags}] The first section  provides further examples of  synthesis  DAGs (directed acyclic graphs) and explains how they differ from trees.
 \item[Section~\ref{sect:furtherModelDetails}] The second section expands upon Section~3 of the main paper by providing further details on our model
   including an algorithm (Alg.~\ref{alg:generator}) showing in detail the generative process.
 \item[Section~\ref{sect:expDetails}] The third section provides further details on our experimental setup.
   For instance it includes details of how we generate our dataset, and how the synthesis score
is calculated. It also contains details of the hyperparameters considered and the computing infrastructure used.
 \item[Section~\ref{sect:expResults}] The fourth section provides further experimental results, such as additional sanity checks for the generation task as well as further figures and tables for the 
optimization tasks. We also provide details of the best molecules found when optimizing for QED and penalized logP, given the popularity of these metrics in previous work.
  \item[Section~\ref{sect:moreRelated}] The fifth and final section delves further into the related work discussed in the introduction
    and related work section of the main paper, giving a more detailed background on the problem that we are tackling and further context on previous approaches.
\end{description}

\section{Further examples of synthesis DAGs}
\label{sect:synDags}

\begin{figure}[h]
\centering
\includegraphics[width=0.99\textwidth]{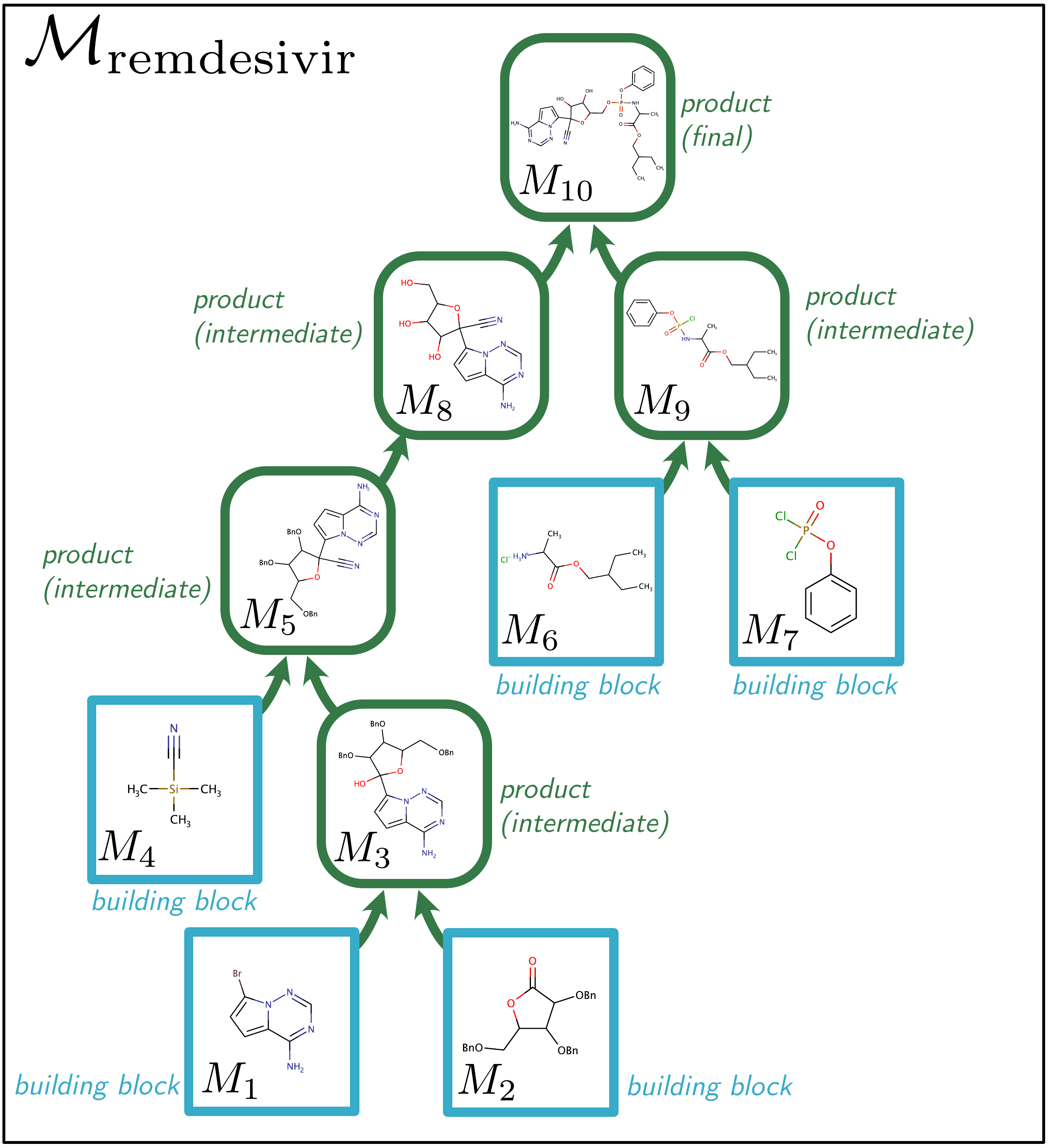}
\caption{
  Simplified synthesis DAG for remdesivir (note ignoring some reagents, conditions and details of chirality) \citep{doi:10.1021/acs.jmedchem.6b01594}. 
  The remdesivir DAG demonstrates a case where two intermediates products need to be formed to then act later as reactants ("Bn" denotes benzyl).}
\label{fig:remdesivir}
\end{figure}

\paragraph{Remdesivir}
We show a simplified synthesis DAG (directed acyclic graph) of Remdesivir \citep{doi:10.1021/acs.jmedchem.6b01594} in Figure~\ref{fig:remdesivir}.
This demonstrates a case where two intermediate products need to be created as reactants for a reaction, and so differs from some of the other simpler
synthesis DAGs presented in our work, which consist of the sequential addition of simple building blocks.

\begin{figure}[h]
\begin{subfigure}{.5\textwidth}
  \centering
  \includegraphics[width=.99\linewidth]{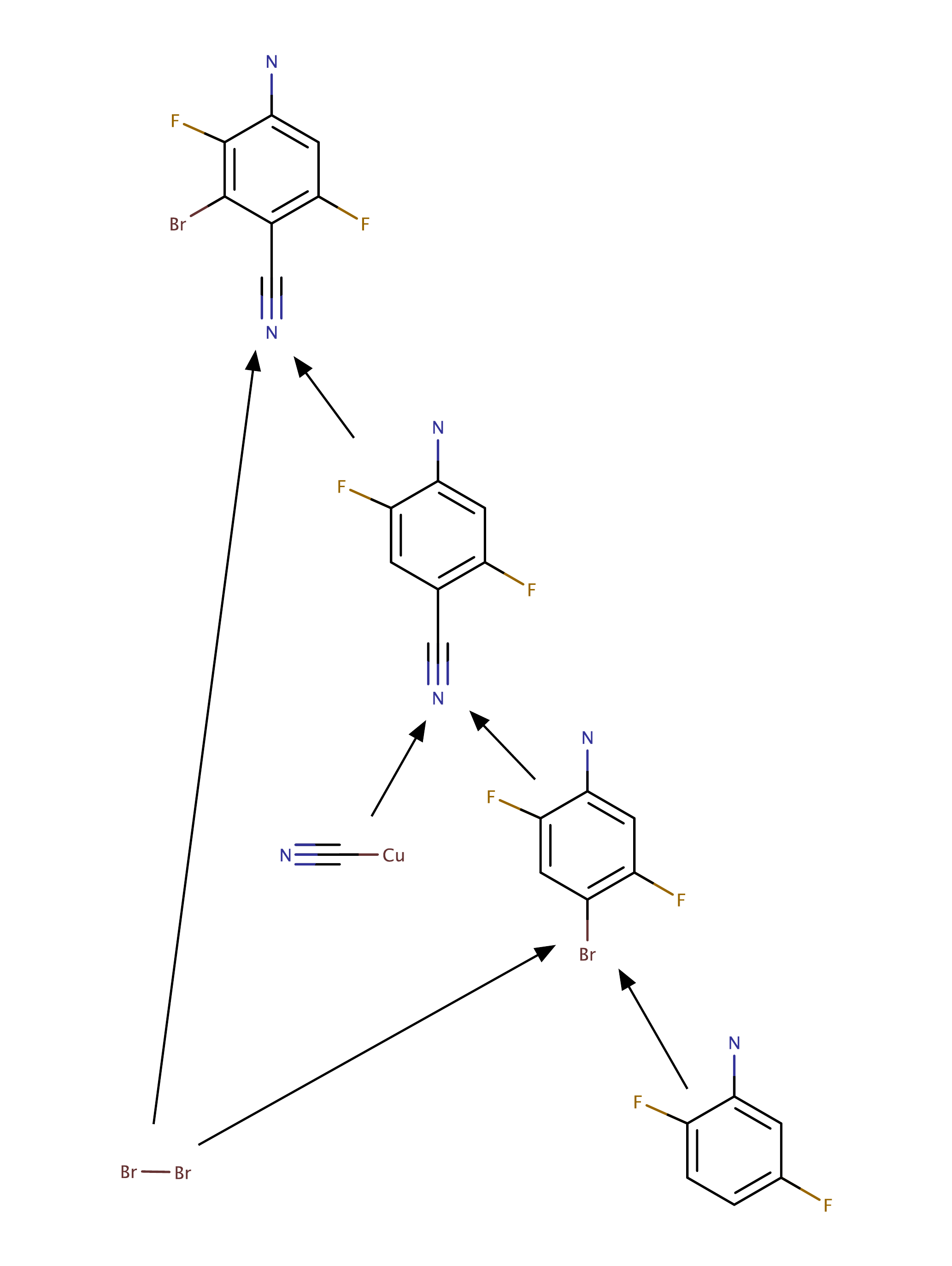}
  \caption{representing using \emph{DAG formalism}}
\end{subfigure}%
\begin{subfigure}{.5\textwidth}
  \centering
  \includegraphics[width=.99\linewidth]{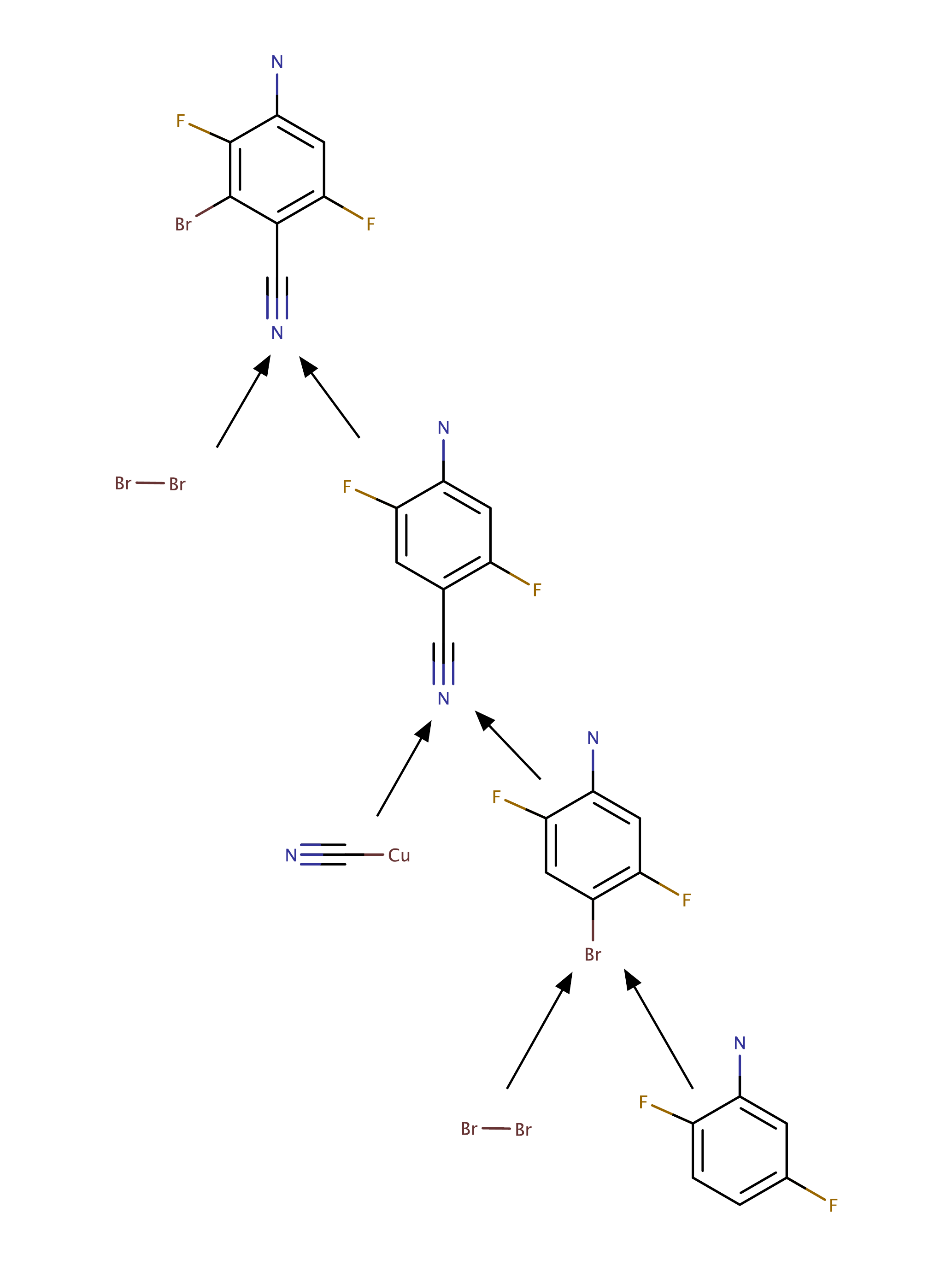}
  \caption{representing using \emph{Tree formalism}}
\end{subfigure}
\caption{Difference between representing the synthesis route as a DAG (with one-to-one mapping between each unique molecule and each node)
or as a tree (a special form of directed acyclic graph where each node can only have one successor). 
Although the representations can easily be converted from one form to another, when generating, if a complex intermediate needs to be reused, 
then it would have to be generated twice in the tree formalism.
}
\label{fig:tree_vs_dag}
\end{figure}

\paragraph{Synthetic routes as DAGs} 
Figure~\ref{fig:tree_vs_dag} shows the difference between representing synthetic routes as a DAG (directed acyclic graph),
as we do in the main paper, and as a tree. 
In the DAG formalism there is a one-to-one mapping between each molecule and each node (see for example \ce{Br2} in the aforementioned figure).
For complete synthetic routes, it is easy to convert from one formalism to another by duplicating or folding nodes into each other.

The advantage of the DAG formalism comes at generation time. During generation, when using the DAG formalism, we only need to generate each unique molecule once. 
This can be advantageous if a complex intermediate product, which requires several steps to create, needs to be used multiple times in the graph,
as in the DAG formalism it only needs to be created once.

\FloatBarrier

\section{Further details about our model}
\label{sect:furtherModelDetails}

In this section we provide further details of our model. 
Our explanation is further broken down into three subsections. 
In the first we provide more details on our generative model for synthesis DAGs, including pseudocode for the full generative process.
In the second subsection we provide further details on how we use a reaction predictor (here: a Molecular Transformer) for reaction prediction to fill in the products of
reactions at test time.
In the third and final subsection we provide further information on the fine-tuning setup.
The description of the hyperparameters and specific architectures used in our models are given in the next section and further details can also 
be found by referring to our code.

\subsection{A generative model of synthesis DAGs}
In this subsection we provide a more thorough description of our generative model for synthesis DAGs.
We first recap the notation that we use in the main paper.
We formally represent the DAG, $\Mc$, as a sequence of actions, with $\Mc = [ V^1,\dots, V^L ]$.
Alongside this we denote the associated action types as $\bm{A} = [A^1, \dots, A^L]$. 
The action type entries $A^l$ take values in $\{ \Aa, \Ab, \Ac \}$, corresponding to the three action types.
The action type entry at a particular step, $A^l$, is fully defined by the actions (and action types) chosen previously to this time $l$,
the exact details of which we shall come back to later.
Finally the set of molecules existing in the DAG at time $l$ are denoted (in an abuse of our notation) by $\MolAtL$.

\paragraph{Actions and the values that they can take} We now describe the potential values that the actions can take. 
These depend on the action type at the step, and we denote this conditioning as $V^l_{|A^l}$.
For example for node addition actions $A^l = \Aa$, the possible values of $V^l$ (i.e. $V^l_{|A_l=\Aa}$) are either \bbnode {} for creating 
a new building block node, or \pnode {} for a new product node.
Building-block actions $A^l = \Ab$ have corresponding values $V^l \in \mathcal{R}$, which determine which building block
becomes a new `leaf' node in the DAG.
Connectivity choice actions $A^l = \Ac$ have values $V^l \in \MolAtL \cup \{ \interProd, \finalProd \} $,
where $\MolAtL$ denotes the current set of all molecules present in the DAG; selecting one of these molecules adds an edge into the new product node. 
The symbol $\interProd$ is an intermediate product stop symbol,
indicating that the new product node has been connected to all its reactants (i.e. an intermediate product has been formed); the symbol $\finalProd $ is a final stop symbol,
which triggers production of the final product and the completion of the generative process.

As hinted at earlier, the action type, $A^l$ is defined by the previous actions $V^1, \dots, V^{l-1}$ and action types (see also Figure 
2 in the main paper). More specifically, this happens as follows:
\begin{itemize}
  \item[] $V^{l-1} =$ \bbnode, then the next action type is building block selection, $A^l=\Ab$.
  \item[]  $A^{l-1} = \Ab$ , then the next action type is again node addition, $A^l=\Aa$ (as you will have selected a building block on the previous step).
  \item[] $V^{l-1} = $ \pnode , then the next action type is connectivity choice, $A^l=\Ac$, to work out what to connect up to the product node previously selected.
  \item[]  $A^{l-1} =\Ac$ then:
    \begin{itemize}
      \item if  $V^{l-1} = \interProd$ then the next action type is to choose a new node again, i.e. $A^l=\Aa$; 
      \item if  $V^{l-1} = \finalProd$ the generation is finished;
    \item if $V^{l-1} \in \MolAtL$ then connectivity choice continues, i.e. $A^l=\Ac$. 
    \end{itemize}
\end{itemize}

\begin{figure}[p]
\centering
\makebox[0pt]{\begin{minipage}{1.1\textwidth}%
\includegraphics[width=1.\textwidth]{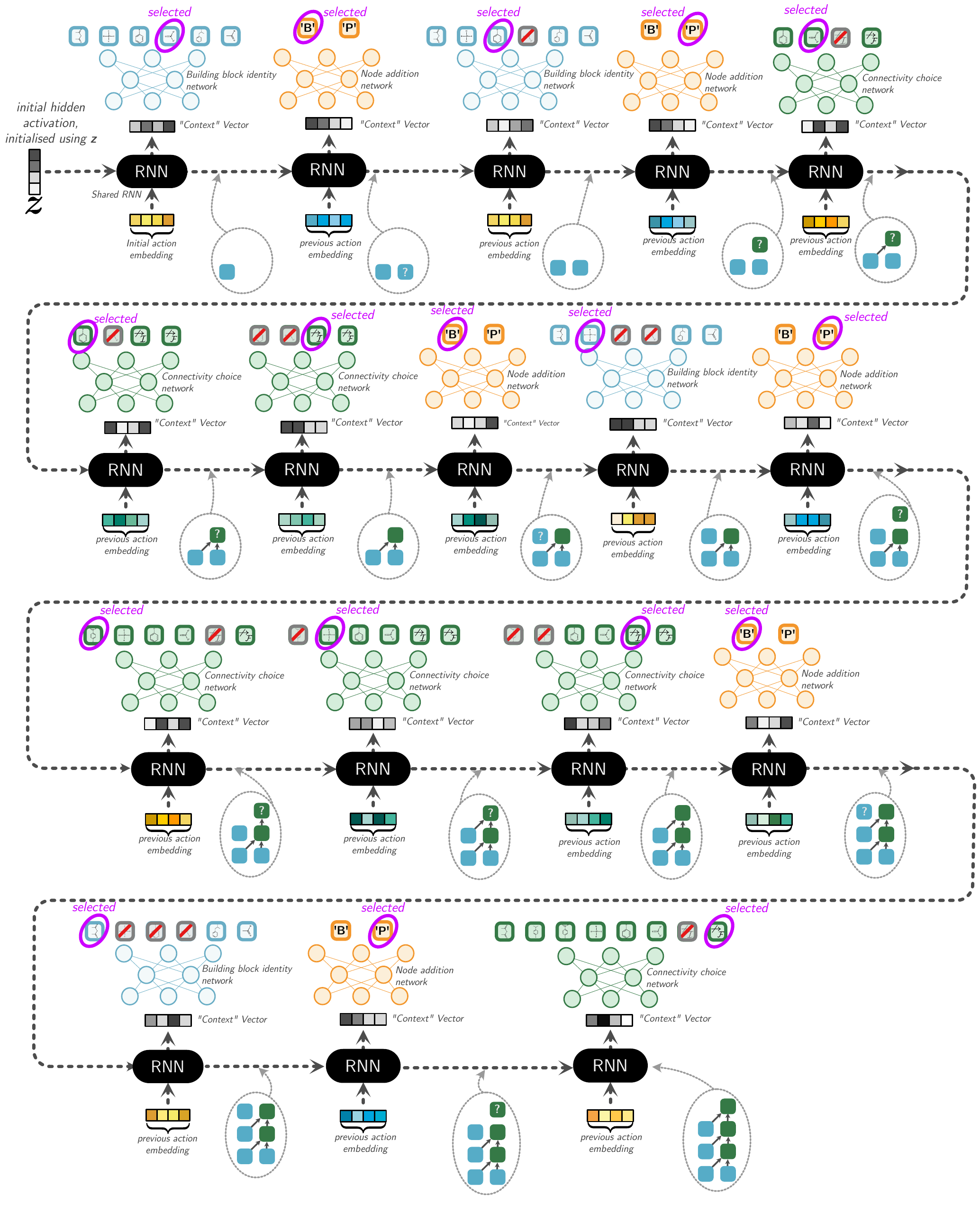}
\caption{
  This is an expanded version of Figure 3 in the main paper showing all the actions required to produce the demonstration, example DAG for paracetamol (see also Figure 1
  and 2 in the main paper). 
  A shared RNN (recurrent neural network) provides a context vector for the different action networks. 
  Based on this context vector, each type of action network chooses an action to take (some actions are masked out as they are not allowed,
  for instance suggesting a 
  building block already in the graph, or choosing to make an intermediate product before choosing at least one reactant).
  Note embeddings of molecular graphs are computed using a GNN (graph neural network).
  The initial hidden vector of the shared RNN is initialized using a latent vector $\bm{z}$ in our autoencoder model DoG-AE; in DoG-Gen it is
  set to a constant.
  The state of the DAG at each stage of the generative process is indicated in the dotted gray circles.
}
\label{fig:reconstruct-actions-long} 
\end{minipage}}
\end{figure}

\paragraph{Our generative process over these actions}
Our model is shown at a high level in Figure~\ref{fig:reconstruct-actions-long} (see also Figure 3 of the main paper), 
which serves to provide an intuitive understanding of the generative process.
The overall structure of the probabilistic model is rather complex, as it depends on a series of branching conditions:
we therefore give pseudocode for the entire generative procedure in detail as a probabilistic program in Algorithm~\ref{alg:generator}.
The program described in Alg.~\ref{alg:generator} defines a distribution 
over DAG serializations; running it forward will sample from the generative process,
but it can equally well be used to evaluate the probability of a DAG $\Mc$ of interest by instead accumulating the log probability
of the sequence at each distribution encountered during the execution of the program.
Note that given our assumption that all reactions are deterministic and produce a single primary product,
product molecules do not appear in our decomposition.

\newcommand{\bB}{\bm{B}}
\newcommand{\be}{\bm{e}}

\begin{figure}[p]
\begin{algorithm}[H]
  \footnotesize
  \caption{Probabilistic simulator for serialized DAGs}
  \label{alg:generator}
  \begin{algorithmic}[1]
  \Require Action networks for node addition, building block molecular identity, and connectivity choice: $\mathsf{na}(\cdot)$,
           $\mathsf{bbmi}(\cdot)$, $\mathsf{cc}(\cdot)$;
  \Require Reaction predictor: $\mathsf{Product}(\cdot)$;
  \Require Context RNN: $\cb^{l} = r(\cb^{l-1}, \bm{e}^l)$;
  \Require Continuous latent variable: $\bm{z}$
  \Require Linear projection for mapping continuous latent to RNN initial hidden: $\mathsf{Lin}(\cdot)$.
  \Require Gated graph neural network, $\textsf{GGNN}(\cdot)$, for computing molecule embeddings
  \Require Learnable embeddings for abstract actions: $\bm{h}_\textrm{\bbnode}$, $\bm{h}_\textrm{\pnode}$, $\bm{h}_{\interProd}$, and 
  $\bm{h}_{\finalProd}$.

  \State Initialize DAG $\Mc \gets [V^1 = \textrm{\bbnode} ]$, Initialize $\bm{A} \gets [A^1=\Aa, A^2=\Ab]$
  \State Initialize molecule set $M \gets \{ \} $ and set of unused reactants $U \gets \{ \}$ \Comment{Track all / all unused molecules}
  \State $\cb^1 \gets \mathsf{Lin}(\bm{z})$  \Comment{Z initializes the first hidden state of the recurrent NN}
  \State $\be^2 \gets  \bm{h}_\textrm{\bbnode}$  \Comment{Initial input into RNN reflects that new node added on first step.}
  \While{ $V^{|\Mc|} ~ \neq ~  \finalProd $}\Comment{Loop until stop symbol}
	  \State $l \gets |\Mc| + 1$
    \State $\cb^l \gets r(\cb^{l-1}, \be^l)$\Comment{Update context}
    %
	  \If{$A^{l} = \Aa$} \Comment{Add a new node}; 
     \State $\bm{w} \gets \mathsf{na}(\cb^l); \quad \bm{\bB} \gets \textrm{STACK}([\bm{h}_\textrm{\bbnode}, \bm{h}_\textrm{\pnode}])$
     \State $\textrm{logits} \gets \bm{w}\bB^T $
     \State $ V^l \sim  \textrm{softmax}(\textrm{logits})$
	  	\If {$V^{l} = $ \bbnode}\Comment{new building block}
      \State $A^{l+1} \gets \Ab; \quad  \be^{l+1} \gets \bm{h}_\textrm{\bbnode}$
		\ElsIf {$V^l = $ \pnode}\Comment{new product}
      \State $A^{l+1} \gets \Ac;  \quad  \be^{l+1} \gets \bm{h}_\textrm{\pnode}$
      \State Initialize intermediate reactant set $R \gets \{ \}$ \Comment{Will temporarily store \emph{working} reactants}
      \State $\textrm{stop\_actions} \gets [\bm{h}_{\finalProd}]$ \Comment{You cannot stop for intermediate product until at least one reactant}
		\EndIf
    %
    \ElsIf {$A^l = \Ab$} \Comment{Pick building block molecular identity}
    \State $\bm{w} \gets \mathsf{bbmi}(\cb^l); \quad \bm{\bB} \gets \textrm{STACK}([\mathsf{GGNN}(g) ~  \textrm{for} ~ g ~ \textrm{in} ~ \mathcal{R} \setminus M])$
    \State $\textrm{logits} \gets \bm{w} \bB^T$
    \State $ V^l \sim  \textrm{softmax}(\textrm{logits})$ \Comment{Pick building block molecule}
    \State $A^{l+1} \gets \Aa; \quad \be^{l+1} \gets \mathsf{GGNN}(V^l)$ 
    \State $M \gets M \cup \{ V^l \}$, $U \gets U \cup \{ V^l \}$
    %
    \ElsIf {$A^l = \Ac$} \Comment{Connectivity choice}
    \State $\bm{w} \gets \mathsf{cc}(\cb^l); \quad \bm{\bB} \gets \textrm{STACK}([\mathsf{GGNN}(g) ~  \textrm{for} ~ g ~ \textrm{in} ~ M \setminus R] +  \textrm{stop\_actions} )$
    \State $\textrm{logits} \gets \bm{w} \bB^T$
    \State $ V^l \sim  \textrm{softmax}(\textrm{logits})$ \Comment{Pick either (i) molecule to connect to, or (ii) to end and create product}
	  \If{$V^l = \interProd$}\Comment{Selected an intermediate product so run the reaction}
	        		\State $M^\mathrm{new} \gets \mathsf{Product}( R )$
        			\State $M \gets M \cup \{ M^\mathrm{new} \}$, $U \gets U \cup  \{ M^\mathrm{new} \}$
              \State $A^{l+1} \gets \Aa; \quad \be^{l+1} \gets \bm{h}_{\interProd}$
    \ElsIf{ $V^l \in M$ } \Comment{Selected an extra reactant}
      \State  $R \gets R \cup \{ V^l \}$ \Comment{Update reactant set}
      \State $U \gets U \setminus \{ V^l \}$\Comment{Remove from pool of ``unused'' molecules}
      \State $A^{l+1} \gets \Ac; \quad  \be^{l+1} \gets \mathsf{GGNN}(V^l)  $ 
      \State $\textrm{stop\_actions} \gets [\bm{h}_{\interProd}, \bm{h}_{\finalProd}]$ \Comment{Now you can stop for both final or intermediate product}
	  \EndIf
    \EndIf
    \State Update $\Mc \gets [ V^1, \dots,  V^l]; \quad \bm{A} \gets [A^1,\dots,A^l] $
  \EndWhile
  \State Predict final product $M_T \gets \mathsf{Product}( R \cup U )$\Comment{The final product considers both $R$ and $U$}
  \State \Return $\Mc$, $M_T$
  \end{algorithmic}
\end{algorithm}
\end{figure}

\subsection{Reaction prediction}
\label{sect:reactionPredictionApp}

As described in the main paper we use the Molecular Transformer \citep{schwaller2018molecular} for reaction prediction.
We use pre-trained weights (trained on a processed USPTO \citep{Jin2017-hh, lowe2012extraction} dataset without reagents). Furthermore, 
we treat the transformer as a black box oracle
and so make no further adjustments to these weights when training our model. Therefore, please note that our algorithm is not restricted to only using the Molecular Transformer ---  any reaction prediction model can be employed with our algorithm, and our algorithm will benefit from the development of stronger reaction predictors in the future.  
We take the top one prediction from the transformer as the prediction for the product, and if this is not a valid molecule (determined by RDKit) 
then we instead pick one of the reactants randomly. 

When running our model at prediction time there is the possibility of getting loops (and so no longer predicting a DAG)
if the output of a reaction (either intermediate or final) creates a molecule which already exists  (in the DAG) as a predecessor of one of the reactants.
A principled approach one could use to deal with this when using a probabilistic reaction predictor model, such as the Molecular Transformer,
is to mask out the prediction of reactions that cause loops in the reaction predictor's beam search.
However, in our experiments we want to keep the reaction predictor as a black box oracle, for which we send reactants and for which it sends us back a product.
Therefore, to deal with any prediction-time loops we go back through the DAG, before and after predicting the final product node, and
remove any loops we have created by choosing the first path that was predicted to each node.

\subsection{Fine-tuning}

The algorithm we use for fine-tuning is given in Algorithm~\ref{alg:fine-tuning}.

\begin{algorithm}[H]
  \footnotesize
  \caption{Synthesis DAG Fine-Tuning. Note that for fine-tuning we use the model DoG-Gen, in which $\bm{z}$ is always set at  $\bm{0}$, hence we drop our specific 
  dependence on $\bm{z}$ in this algorithm.}
  \label{alg:fine-tuning}
  \begin{algorithmic}[1]
    \Require Initial model $p_\theta(\Mc)$, iterations $I$, threshold $K$, sample size $N$, objective $h(\cdot)$, pool of seen synthesis DAGs (initially empty), $P$.
 	  \State Compute the score $h(\cdot)$ of all products in the initial training set and add to $P$
    \For{$i = 1,\dots, I$}
	  \State Sample $N$ DAGs from $p_\theta(\Mc)$ 
	  \State Compute the score $h(\cdot)$ of the $N$ products and add to pool, $P$
	  \State Select the $K$ DAGs with the highest score from $P$
	  \State Run two training epochs on $\theta$ using these $K$ DAGs as training data
  \EndFor
  \State \Return updated model $p_\theta(\Mc)$, pool of all seen synthesis DAGs (ranked) $P$
  \end{algorithmic}
\end{algorithm}

\section{Further experimental details}
\label{sect:expDetails}

This section provides further details about aspects of our experiments.
We start by describing how we create a dataset of synthesis DAGs for training. We then describe how the synthesis score we use in the optimization experiments is calculated.
Finally, the latter subsections provide  specific details on the hyperparameters we use. 
Further details about our experiments can also be found in our code.

\subsection{Creating a dataset of synthesis DAGs}
\label{sect:DatasetCreation}

\begin{figure}[t!]
\centering
\includegraphics[width=\textwidth]{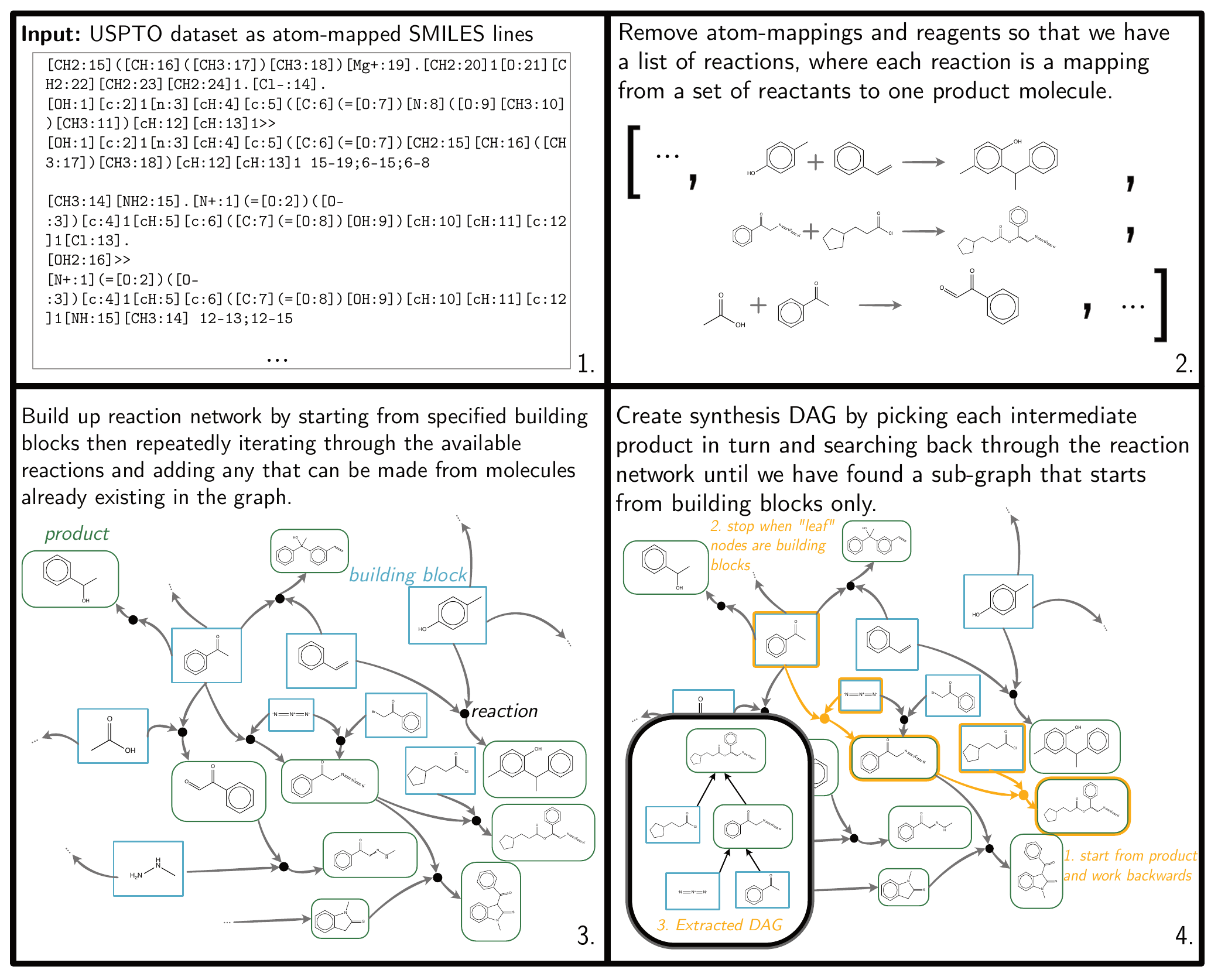}
\caption{
An illustration of how we create a dataset of synthesis DAGs from a dataset of reactions. 
We first clean up the reaction dataset by removing reagents (molecules which do not contribute atoms to the final product) 
and any reactions which lead to more than one product.
We then form a modified reaction network (we do not allow loops back to building block molecules), which is a directed graph showing how molecules are linked to others through reactions.
This process starts by adding molecule nodes corresponding to our initial building blocks.
We then repeatedly iterate through our list of reactions and gradually add reaction nodes (and their associated product nodes)
to the graph if both (i) the corresponding reaction's reactants are a subset of the molecule nodes already in the graph, and (ii) 
the product is not a building block.
Finally for each possible product node we iterate back through the directed edges until we have selected a subgraph without any loops,
where the initial nodes are members of our set of building blocks.
}
\label{fig:createingDataset}
\end{figure}

In this subsection we describe how we create a dataset of synthesis DAGs, with a high level illustration of the process given
in Figure~\ref{fig:createingDataset}. Further details can also be found by referring to our code.
The creation of our synthesis DAG dataset starts by collecting the reactions from the USPTO dataset \citep{lowe2012extraction}, 
using the processed and cleaned version of this dataset provided by \citep[\S4]{Jin2017-hh}.
We filter out reagents (molecules that do not contribute any atoms to the final product) and multiple product reactions 
(97\% of the dataset is already single product reactions)
using the approach of \citet[\S 3.1]{schwaller2018found}.

This processed reaction data is then used to create a reaction network \citep{segler2017modelling,Jacob2018-bx, Grzybowski2009-ti}.
To be more specific, we start from the reactant building blocks specified in \citet[\S 4]{bradshaw2019model} as
initial molecule nodes in our network,
and then iterate through our list of processed reactions adding any reactions (and the associated product molecules)
(i) that depend only on molecule nodes that are already in our network, and (ii) where the product is not an initial building block.
This process repeats until we can no longer add any of our remaining reactions.

This reaction network is then used to create one synthesis DAG for each molecule. To this end, starting from each possible (non building block) molecule
node in our reaction network, we step backwards through the network 
until we find a sub-graph of the reaction network (without any loops) with initial nodes that are from our collection of building blocks. 
When there are multiple possible routes we pick one.
This leaves us with a dataset of 72008 synthesis DAGs, which we use approximately 90\% of as training data and split the remainder into 
a validation dataset (of 3601 synthesis DAGs) and test dataset (of 3599 synthesis DAGs).

The training set DAGs have an average of 4.6 nodes, with the final molecules containing an average of 20.5 heavy atoms and 21.7 bonds (between heavy atoms).
The average number of actions required to construct these DAGs is 11, as each reactant often contributes several atoms and bonds to the product.

\subsection{Synthesizability Score}
\label{sect:appSynScore}

The synthesizability score is defined as the geometric mean of the nearest neighbor reaction similarities:
\begin{equation}
\sqrt[|R|]{\prod_{ r \in R } \kappa(r, \text{nn}(r))}
\end{equation}
where $R$ is the list of reactions making up a synthesis DAG, $ r \in R$ are the individual reactions in the DAG, $\text{nn}(r)$ is the nearest neighbor
reaction in the chemical literature in Morgan fingerprint space,
and $\kappa(\cdot, \cdot)$ is Tanimoto similarity over Morgan reaction fingerprints \cite{schneider2015development}.

\subsection{Atom features used in DoG models}

The atom features we use as input to our graph neural networks (GNNs) operating on molecules are given in Table~\ref{table:atom-features}.
These features are chosen as they are used in \citet[Table 1]{gilmer2017neural}
(we make the addition of an expanded one-hot atom type feature, to cover the greater range of elements present in our molecules).

\begin{table}[H]
  \caption{Atom features we use as input to the GGNN. These are calculated using RDKit.}
  \label{table:atom-features}
  \centering
  \begin{tabular}{ll}
    \toprule
    Feature     & Description      \\
    \midrule
    Atom type & 72 possible elements in total, one hot  \\
    Atomic number & integer \\
    Acceptor & boolean (accepts electrons) \\
    Donor & boolean (donates electrons) \\
    Hybridization & One hot (SP, SP2, SP3) \\
    Part of an aromatic ring & boolean\\
    H count & integer \\
    \bottomrule
  \end{tabular}
\end{table}

\subsection{Implementation details for DoG-AE}
\label{sect:implementationDetailsAE}

In this subsection we describe specifics of our DoG-AE model used to produce the results in Table~1 of the main paper.

\paragraph{Forming molecule embeddings} For forming molecule embeddings we use a GGNN (Gated Graph Neural Network) \citep{li2015gated}; this operates on the 
atom features described in Table \ref{table:atom-features}. This graph neural network (GNN) was run for 4 propagation steps to update the node embeddings,
before these embeddings were projected down to a 50 dimensional space using a learnt linear projection. The node embeddings were then combined to form
molecule embeddings through a weighted sum. The same GNN architecture was shared between the encoder and the decoder.

\paragraph{Encoder} 
The encoder consists of two GGNNs. The first, described above, creates molecule embeddings which are then used to initialize the node embeddings in the synthesis DAG.
The synthesis DAG node embeddings, which are 50 dimensional, are further updated using a second GGNN. Using the notation introduced in Section~3.2.1 of the main paper, we have:
\begin{align}
   g_e(\mb_M^t, \eb_M^{t-1}) = \textrm{GRU}(\mb_M^t, \eb_M^{t-1}) \\
   g_m(\eb_M^{t-1}, \eb_{N(M)}^{t-1}) = \sum_{j \in N(M)} \textrm{Lin}_1(\eb_j^{t-1})
\end{align}
where $\textrm{GRU}(\cdot)$ is a gated recurrent unit (GRU) \citep{cho-etal-2014-learning} and $\textrm{Lin}_1(\cdot)$ is a learnt linear projection.
Seven propagation steps of message passing are carried out on the DAG (i.e. $T'=7$), and the messages are passed forward on the DAG from the `leaf' nodes
to the final product node. 
Finally, the node embedding of the final product molecule node in the DAG is passed through an additional linear projection to
parameterize the mean and log variance of independent Gaussian distributions over each dimension of the latent variable, $\bm{z}$, i.e.:
\begin{align}
     g_\mu(\Eb_{\Mc}^{T'})  = \textrm{Lin}_2 \left(  [ \Eb_{\Mc}^{T'}  ]_{M^T}  \right) \\
     g_{\log \sigma^2}(\Eb_{\Mc}^{T'})  = \textrm{Lin}_3 \left(  [ \Eb_{\Mc}^{T'}  ]_{M^T}  \right) 
\end{align}
where we are using the notation $[\Eb_{\Mc}^{T'} ]_{M^T}$ to indicate the indexing of the node embedding corresponding to the final product node ($M^T$) in the DAG ($\Mc$),
after $T'$ stages of message passing.

\paragraph{Decoder}
For the decoder we use a 3 layer GRU RNN \citep{cho-etal-2014-learning} to compute the context vector. 
The hidden layers have a dimension of 200 and whilst training we use a dropout rate of $0.1$.
For initializing the hidden layers of the RNN we use a linear projection (the parameters of which we learn) of $z$.
The action networks are feedforward neural networks with one hidden layer (dimension 28) and ReLU activation functions.
For the abstract actions (such as \bbnode or \pnode) we learn 50 dimensional embeddings, such that these embeddings have the same dimensionality as 
the molecule embeddings we compute.

\paragraph{Training} 
We train our model, with a 25 dimensional latent space, using the Adam optimizer \citep{kingma2014adam}, an initial learning rate of 0.001, and a batch size of 64. 
We train the autoencoder using the Wasserstein autoencoder loss
\citep{tolstikhin2017wasserstein} (Eq. 2 main paper), with $\lambda=10$ and an inverse multiquadratics kernel for computing the MMD-based penalty,
as this is what is used in \citet[\S 4]{tolstikhin2017wasserstein}. 

Our model, DoG-AE, is trained using teacher forcing for 400 epochs (each epoch took approximately 7 minutes) and we multiplied the learning rate by a factor of 0.1 after 300 and 350 epochs.
DoG-AE obtains a reconstruction accuracy (on our held out test set) of 65\% when greedily decoding (greedy in the sense of picking the most probable action at each stage of decoding). If we try instead decoding by sampling 100 times from our model and then sorting based on probability we obtain a slightly improved reconstruction accuracy of 66\%.

Initially in our experiments we trained for a shorter period, but we increased the training time after monitoring the reconstruction
rate on our held out validation dataset. 
Likewise for debugging purposes we initially trained with a smaller architecture. To be more specific, we tried
a two layer GRU RNN in our decoder with 28 dimensional hidden layers. We also set the molecule embedding sizes to be 28 dimensional.
We increased the size of our architecture after we found that it improved results. 
However, we have yet to attempt a thorough hyperparameter search which may improve results further.

\paragraph{Timings}
Using a NVIDIA K80 GPU it takes $\approx$ 7 mins to run a training epoch for DoG-AE
($\approx$ 0.4 secs per batch of 64).
At inference time,  where we do not initially have access to the complete sequence, we usually run larger batches, due to the fixed costs and latency of
communicating with a Molecular Transformer server. It takes $\approx$ 29 secs to carry out \emph{per batch of 200 DAGs} (and $\approx$ 12 secs per batch of 64 DAGs).
Our approach could be further sped up by using faster reaction predictors.

\subsection{Implementation details for DoG-Gen}

For DoG-Gen we also used a GGNN to create molecule embeddings in a similar way to DoG-AE. The GGNN was run for 5 rounds of message passing
to form 80 dimensional node embeddings;
these node embeddings were aggregated into a 160 dimensional molecule embedding through a linear projection and weighted sum. 
For generating the context vector we use a 3 layer GRU RNN with 512 dimensional hidden layers.
The action networks  used were feed-forward neural networks with one hidden layer of dimension 28 and ReLU activation functions.
We trained our model for 30 epochs, as we found using our validation dataset, that training for longer would lead to overfitting.
Aside from monitoring the training time we have not tried tuning other hyperparameters,  which we believe could lead to better results.

For optimization we start by evaluating the score on every synthesis DAG in our training and validation datasets;
we then run 30 stages of fine-tuning, sampling 7000 synthesis DAGs at each stage and updating the weights of our model
using the best 1500 DAGs seen at that point as a fine-tuning dataset.
For both our model and the baselines, when reporting the GuacaMol benchmark property score we report
the score obtained by the best individual molecule in the group.

To parallelize over the GuacaMol optimization tasks we used Jug \citep{coelho2017jug}.

\subsection{Details of baselines for generation tasks}

We used the following implementations for the baselines:
\begin{itemize}
\item SMILES LSTM \citep{segler2018generating}: \url{https://github.com/benevolentAI/guacamol_baselines}.
\item JT-VAE \citep{Jin2018-aa}: \url{https://github.com/wengong-jin/icml18-jtnn} (we used the updated version of their code, ie the \texttt{fast\_jtnn} version)
\item CGVAE \citep{Liu2018-ha}: \url{https://github.com/Microsoft/constrained-graph-variational-autoencoder}
\item Molecule Chef \citep{bradshaw2019model}: \url{https://github.com/john-bradshaw/molecule-chef}
\end{itemize}

For the CVAE, GVAE and GraphVAE baselines we used our own implementations.
We tuned the hyperparameters of these models on the ZINC or QM9 datasets so that we were able to get at least similar
(and often better) results compared to those originally reported in \citet{Kusner2017-ry,Simonovsky2018-md}.

When training the GraphVAE on our datasets we exclude any molecules with greater than 20 heavy atoms,
as this procedure was found in the original paper to give better performance when training on ZINC \citep[\S 4.3]{Simonovsky2018-md}.
We use a 40 dimensional latent space, a GGNN \citep{li2015gated} for the encoder, and use max-pooling graph matching during training.

For the CVAE and GVAE we use 72 dimensional latent spaces. 
We multiply the KL term in the VAE loss by a parameter $\beta$ \citep{higgins2017beta,alemi2018fixing};
this $\beta$ term is then gradually annealed in during training until it reaches a final value of $0.3$.
We use a 3 layer GRU RNN \citep{cho-etal-2014-learning} for the decoder with 384 dimensional hidden layers. 
The encoder is a 3 layer bidirectional GRU RNN also with 384 dimensional hidden layers.

\subsection{Computing infrastructure used}

We mostly used a NVIDIA Tesla K80 GPU when training and sampling from our models (predominately using NC6 and NC12 virtual machines on Azure).
At sampling time we ran a Molecular Transformer in server mode on a separate K80 GPU.

For training the baselines for the generation task we used a mixture of NVIDIA Tesla K80, GeForce GTX 1080 Ti and P100 GPUs.
The NVIDIA P100 GPU was used for training the CGVAE as this model required a GPU with a large memory.

\FloatBarrier
\section{Further experimental results}
\label{sect:expResults}

\subsection{Generation -- Additional Sanity Checks}
\begin{figure}[h]
\vspace{-2ex}
\centering
\makebox[0pt]{\begin{minipage}{1.3\textwidth}%
\includegraphics[width=1.05\textwidth]{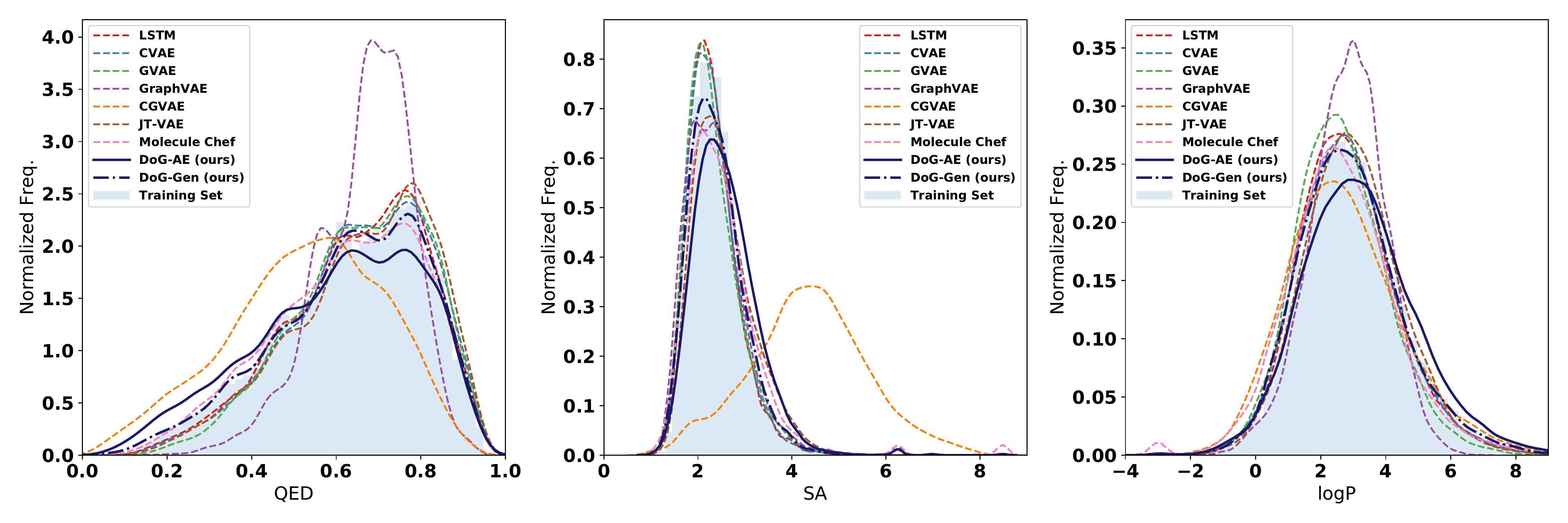}
\vspace{-3ex}
\caption{
KDE (kernel density estimation) plots for the distribution of drug likeness score (QED)
\citep{Bickerton2012-zj}, synthetic accessibility score (SA) \citep{Ertl2009-kj}, and the octanol-water partition coefficient (logP) for 20k
molecules sampled from each of the various models, compared to the training set.
Plot done in same style as \citep[Fig.3]{Seff2019-dk}
}
\label{fig:property-plots}
\end{minipage}}
\end{figure}

Following prior work \citep{Seff2019-dk, polykovskiy2018molecular}, we also plot the distributions of the Quantitative Estimate of Drug Likeness score (QED)
\citep{Bickerton2012-zj}, Synthetic Accessibility score (SA) \citep{Ertl2009-kj}, and the octanol-water partition coefficient (logP)
\citep{Wildman1999-zu} over the generated molecules. The results of this are shown in Figure \ref{fig:property-plots}.
Apart from the GraphVAE and CGVAE, all models closely match the training set. We again see this as a useful model sanity check.

\FloatBarrier

\subsection{Optimization}

\begin{figure}[h]
\centering
\makebox[0pt]{\begin{minipage}{1.3\textwidth}%
\includegraphics[width=\textwidth]{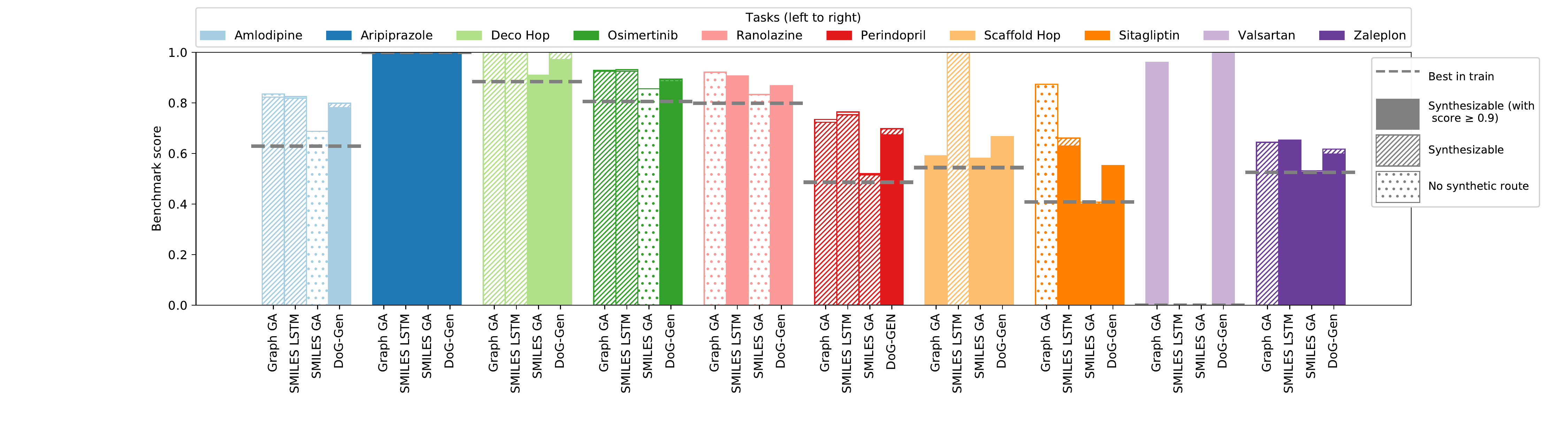}
\caption{
The score of the best molecule found by the different approaches over a series of ten GuacaMol tasks \citep[\S 3.2]{brown2019guacamol}.
Scores range between 0 and 1, with 1 being the best. 
We also differentiate between the synthesizability of the different best molecules found by the hatching on the bar,
differentiating between three different cases: (1) the best molecule found (regardless of synthesizability) is shown with a circular hatched bar,
(2) the best molecule found with a synthetic route with the diagonally hatched bar, and
 (3) the best molecule found which has a synthetic score greater than or equal to 0.9 is given a solid bar.
 Note that later bars occlude previous bars.
The dotted gray line for each task represents the score one obtains if picking the best molecule in the training set.
}
\label{fig:benchmark_scores_thresh09}
\end{minipage}}
\end{figure}

\begin{figure}[h]
\centering
\makebox[0pt]{\begin{minipage}{1.3\textwidth}%
\includegraphics[width=\textwidth]{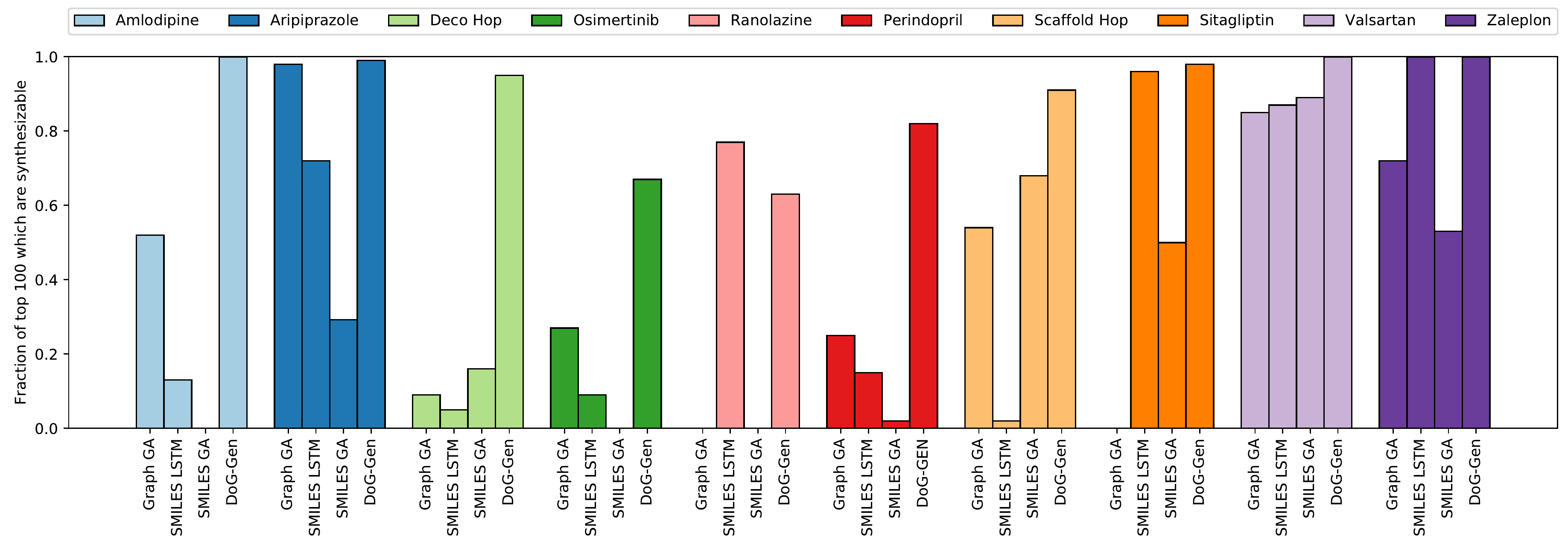}
\caption{
  The fraction of the top 100 molecules proposed that for which a synthetic route can be found, over a series of ten Guacamol tasks \citep[\S 3.2]{brown2019guacamol}.
}
\label{fig:frac_syn}
\end{minipage}}
\end{figure}

\paragraph{Further plots and tables for the GuacaMol optimization tasks}
We provide further figures and tables for the GuacaMol optimization tasks.
Figure~\ref{fig:benchmark_scores_thresh09} shows the scores of the best molecules found on each of the tasks by the various methods,
particularly distinguishing between the cases in which (a) no synthetic route can be found, (b) a synthetic route can be found, 
and (c) a synthetic route can be found \emph{and} the found route has a synthesis score over 0.9.
Figure~\ref{fig:frac_syn} and Table \ref{table:frac_syn} show the fraction of the top 100 molecules proposed by each method for each task
for which a synthetic route can be found.
Table~\ref{table:syn_score} shows the average synthesis score over the 100 best molecules proposed by each method for each task.
Table~\ref{table:median_steps} shows the median number of synthesis steps over the top 100 best molecules for each method and task,
where the median has been calculated considering only those molecules in the top 100 for which a synthetic route can be found.

As an additional baseline, we reimplemented the SYNOPSIS algorithm, one of the first reaction-driven de novo design algorithms based on discrete optimization (simulated annealing), proposed by \citet{vinkers2003synopsis} in 2003. 
The algorithm proceeds by sampling molecules from a pool (which is initially populated by all building blocks and training molecules that the other models also have access to). Then, at each iteration three molecules $m_i$ are sampled according to a Boltzmann distribution $p(m_i) = \frac{ \exp( -(V_{\text{max}} - V_{m_i})/T) }{\sum_j \exp( -(V_{\text{max}} - V_{m_j})/T)}$, with molecules $m_i$, the score of a molecule $V_{m_i}$, the score of the currest best molecule collected so far $V_{\text{max}}$, and a quasi-temperature $T$. 
For these molecules, a reaction scheme is randomly sampled, using a union of the reaction schemes published in \cite{chevillard2015scubidoo,button2019automated}. If the reaction requires a reaction partner, up to 64 building blocks are uniformly-randomly sampled from all matching building blocks. 
The resulting molecules are scored, and added back to the pool and to an output collection, from which the results are eventually returned.
The algorithm is run for 1000 iterations, generating $\approx$ 100,000 molecules per task. 
Table~\ref{table:synopsis} shows the best molecule scores on the GuacaMol tasks found by this reimplementation of \citet{vinkers2003synopsis}'s algorithm in comparison to the DoG-Gen algorithm. 
The results indicate that DoG-Gen also performs favorably  in comparison to an established reaction-driven de novo design algorithm.

\begin{figure}[h]
\centering
\makebox[0pt]{\begin{minipage}{1.3\textwidth}%
\includegraphics[width=1\textwidth]{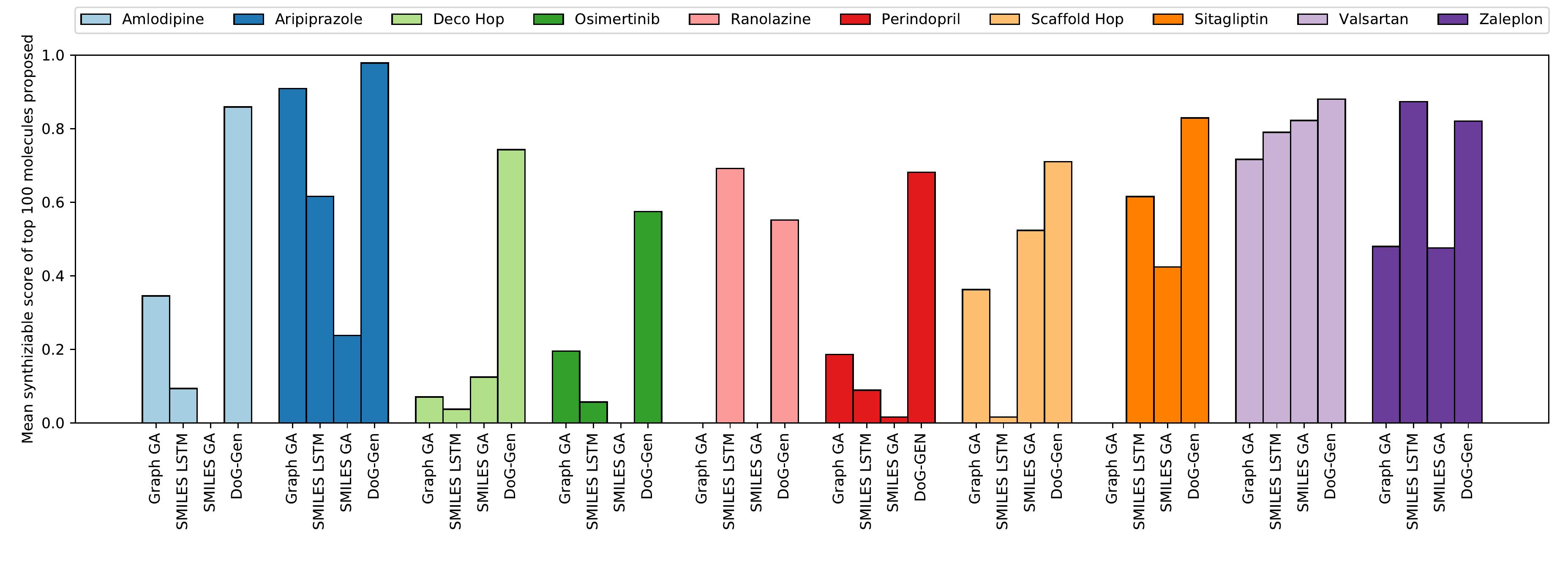}
\caption{
  The mean of the synthesis score of the top 100 molecules proposed by each method, over a series of ten GuacaMol tasks \citep[\S 3.2]{brown2019guacamol}.
}
\label{fig:syn_score}
\end{minipage}}
\end{figure}

\begin{table}[h]
  \begin{center}
    \caption{
      Fraction of the top 100 molecules suggested by each method for which a synthetic route can be found.
    }
\begin{tabular}{lrrrr}
\toprule
 &  Graph GA &  SMILES LSTM &  SMILES GA &  DoG-Gen \\
\midrule
Amlodipine   &      0.52 &         0.13 &       0.00 &     1.00 \\
Aripiprazole &      0.98 &         0.72 &       0.29 &     0.99 \\
Deco Hop     &      0.09 &         0.05 &       0.16 &     0.95 \\
Osimertinib  &      0.27 &         0.09 &       0.00 &     0.67 \\
Perindopril  &      0.25 &         0.15 &       0.02 &     0.82 \\
Ranolazine   &      0.00 &         0.77 &       0.00 &     0.63 \\
Scaffold Hop &      0.54 &         0.02 &       0.68 &     0.91 \\
Sitagliptin  &      0.00 &         0.96 &       0.50 &     0.98 \\
Valsartan    &      0.85 &         0.87 &       0.89 &     1.00 \\
Zaleplon     &      0.72 &         1.00 &       0.53 &     1.00 \\
\bottomrule
\end{tabular}
\label{table:frac_syn}
  \end{center}
\end{table}

\begin{table}[h]
  \begin{center}
    \caption{
      Average synthetic score of the top 100 molecules suggested by each method.
    }
\begin{tabular}{lrrrr}
\toprule
 &  Graph GA &  SMILES LSTM &  SMILES GA &  DoG-Gen \\
\midrule
Amlodipine   &      0.35 &         0.09 &       0.00 &     0.86 \\
Aripiprazole &      0.91 &         0.62 &       0.24 &     0.98 \\
Deco Hop     &      0.07 &         0.04 &       0.12 &     0.74 \\
Osimertinib  &      0.20 &         0.06 &       0.00 &     0.57 \\
Perindopril  &      0.19 &         0.09 &       0.02 &     0.68 \\
Ranolazine   &      0.00 &         0.69 &       0.00 &     0.55 \\
Scaffold Hop &      0.36 &         0.02 &       0.52 &     0.71 \\
Sitagliptin  &      0.00 &         0.62 &       0.42 &     0.83 \\
Valsartan    &      0.72 &         0.79 &       0.82 &     0.88 \\
Zaleplon     &      0.48 &         0.87 &       0.48 &     0.82 \\
\bottomrule
\end{tabular}
\label{table:syn_score}
  \end{center}
\end{table}

\begin{table}[h]
  \begin{center}
    \caption{
      Table showing the median number of synthetic steps for each molecule found in the top 100 (median calculated only over synthesizable molecules).
      Hyphens (`-') indicate no synthesizable molecule was suggested in the top 100 by that method on that task.
    }
\begin{tabular}{lrrrr}
\toprule
          &  Graph GA &  SMILES LSTM &  SMILES GA &  DoG-Gen \\
\midrule
Amlodipine   &       7.0 &          6.0 &        - &      3.5 \\
Aripiprazole &       3.0 &          6.0 &        5.0 &      2.0 \\
Deco Hop     &       5.0 &          8.0 &        3.5 &      5.0 \\
Osimertinib  &       8.0 &          9.0 &        - &      7.0 \\
Perindopril  &       7.0 &         11.0 &        6.0 &      7.0 \\
Ranolazine   &       - &          7.0 &        - &      8.0 \\
Scaffold Hop &       7.0 &          8.0 &        5.0 &      4.0 \\
Sitagliptin  &       - &          7.0 &        3.0 &      3.0 \\
Valsartan    &       6.0 &          3.0 &        2.0 &      3.0 \\
Zaleplon     &       6.0 &          3.0 &        4.0 &      3.0 \\
\bottomrule
\end{tabular}
\label{table:median_steps}
  \end{center}
\end{table}

\begin{table}[h]
  \begin{center}
    \caption{
      Score of the best molecule found by DoG-Gen and the SYNOPSIS \citep{vinkers2003synopsis} inspired, discrete optimization algorithm on the GuacaMol tasks}
\begin{tabular}{lrrrr}
\toprule
 &  DoG-Gen & SYNOPSIS  \\
\midrule
Amlodipine   &    0.80        &  0.63    \\
Aripiprazole &    1.00        &  0.87    \\
Deco Hop     &    1.00        &  0.88    \\
Osimertinib  &    0.89        &  0.84    \\
Perindopril  &    0.70        &  0.55    \\
Ranolazine   &    0.87        &  0.83    \\
Scaffold Hop &    0.67        &  0.54    \\
Sitagliptin  &    0.55        &  0.43    \\
Valsartan    &    1.00        &  0.00    \\
Zaleplon     &    0.62        &  0.52    \\
\bottomrule
\end{tabular}
\label{table:synopsis}
  \end{center}
\end{table}

\FloatBarrier

\paragraph{QED and penalized logP Optimization}
We also considered optimizing for QED \citep{Bickerton2012-zj} and penalized logP\footnote{
  Using the ZINC-normalized definition of this score, as computed for instance in the code of \citet{you2018graph} at 
\url{https://github.com/bowenliu16/rl_graph_generation/blob/master/gym-molecule/gym_molecule/envs/molecule.py}.
}, as these metrics have been used in previous work \citep{you2018graph,Jin2018-aa,Kusner2017-ry}.
We did not include this evaluation in the main paper, as similar to \citet[\S 5.2 \& \S 8.5]{brown2019guacamol} we found that our approach 
was able to obtain very high scoring molecules quite easily, which meant that this score was less useful in investigating the advantages and disadvantages of each method.
However, for reference we note that DoG-Gen found a molecule with top QED score of 0.948%
\footnote{With SMILES: \texttt{Cc1cc(F)ccc1NS(=O)(=O)c1ccc2c(c1)CCO2}} and a top penalized
logP score of 124.8%
\footnote{With SMILES: 
  \texttt{CCCCCCCCCCCCCCCCCCCCCCCCCCCCCCCCCCCCCCCCCCCCCCCCCCCCCCCCCCCCCCCCCCCCCCCCCCCCCCCCCC\\
    CCCCCCCCCCCCCCCCCCCCCCCCCCCCCCCCCCCCCCCCCCCCCCCCCCCCCCCCCCCCCCCCCCCCCCCCCCCCCCCCCCCCCCCCCCCCCCCCC\\
    CCCCCCCCCCCCCCCCCCCCCCCCCCCCCCCCCCCCCCCCCCCCCCCCCCCCCCCCCCCCCCCCCCCCCCCCCCCCCCCCCCCCCCCCCCCCCCCCC\\
    CCCCCCCCCCCCCCCCCCCCCCCCCCCCCCCCCCCCCCCCCCCCCCCCCCCCCCCCCCCCCCCCCCCCCCCCCCCCCCCCCCCCCCCCCCCCCCCCC\\
    CCCCCCCCCCCCCCCCCCCCCCCCCCCCCCCCCCCCCCCCCCCCCCCCCCCCCCCCCCCCCCCCCCCCCCCCCCCC[S+](CCCCCCCCCCCCCCCC\\
CC)CCCCCCCCCCCCCCCCCCCCCCCCCCCCCC}}
(we run the penalized logP optimization for only 15 rounds of fine-tuning as it quickly exploits the objective by producing a molecule with a
long carbon chain.).

\subsection{Retrosynthesis}

Our generative model of synthesis DAGs can be seen as a parametrizable mapping from a vector of real numbers to a synthesis DAG.
As a module it can be mixed and matched in different ML frameworks as we have already seen in the main paper with DoG-AE and DoG-Gen.
In this section we describe some preliminary results with a third model architecture, called RetroDoG. 
This model consists of the composition of a GNN followed by our generative synthesis DAG model to produce a learnable
mapping from a molecular graph to a synthesis DAG. By training this model on pairs of product molecules 
and their associated synthesis DAGs we can use this model to perform retrosynthesis (i.e. predict how a particular product can be made). 
While automated retrosynthesis is canonically performed using planning algorithms \cite{segler2018planning}, the model described in this section would additionally allow one to feed in a potentially hard or impossible to synthesize molecule, and obtain a similar molecule which is \textit{easy to synthesize}, which is impossible with current planners.

\begin{figure}[h]
\centering
\includegraphics[width=\textwidth]{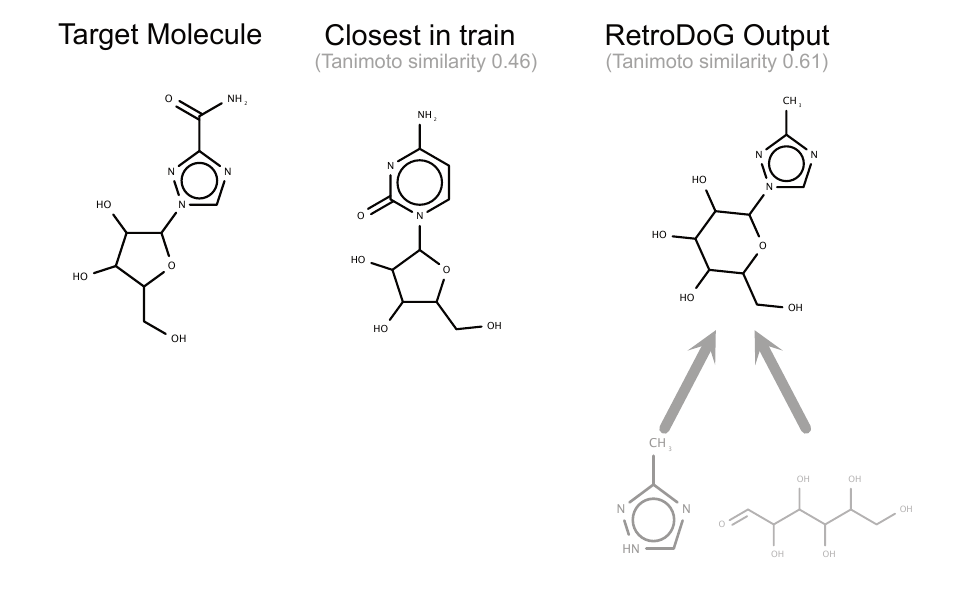}
\caption{
RetroDoG suggested DAG for ribavirin.
}
\label{fig:ribavirin}
\end{figure}

\begin{figure}[h]
\centering
\includegraphics[width=\textwidth]{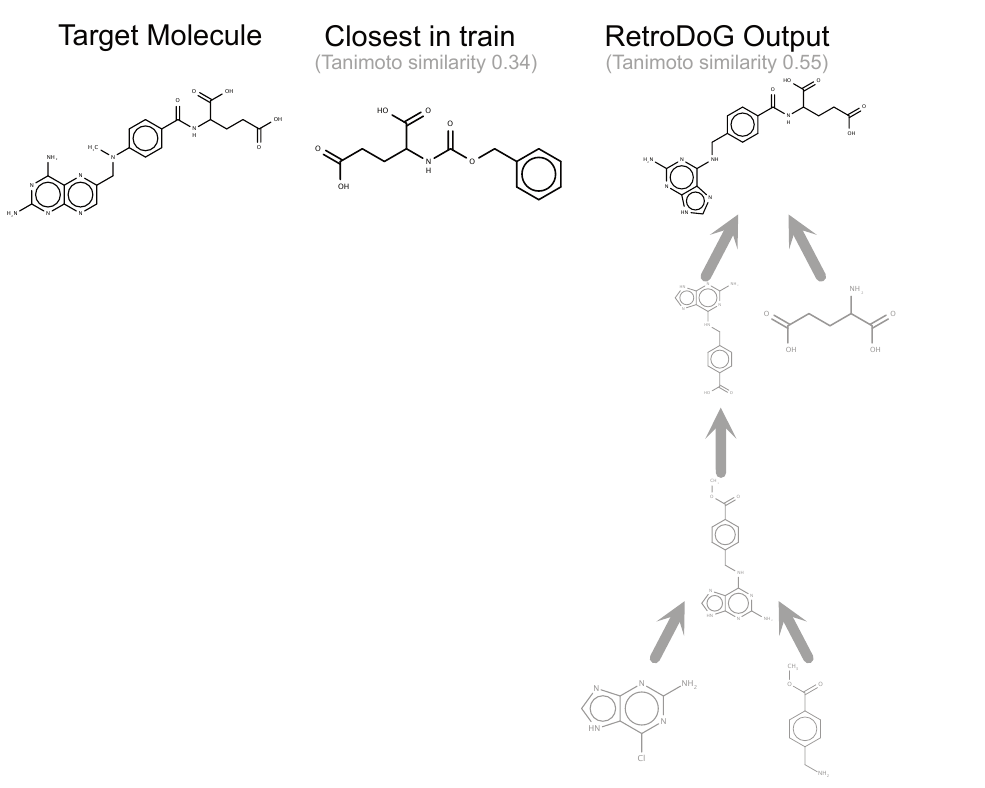}
\caption{
RetroDoG suggested DAG for methotrexate.
}
\label{fig:methotrexate}
\end{figure}

\begin{figure}[h]
\centering
\includegraphics[width=\textwidth]{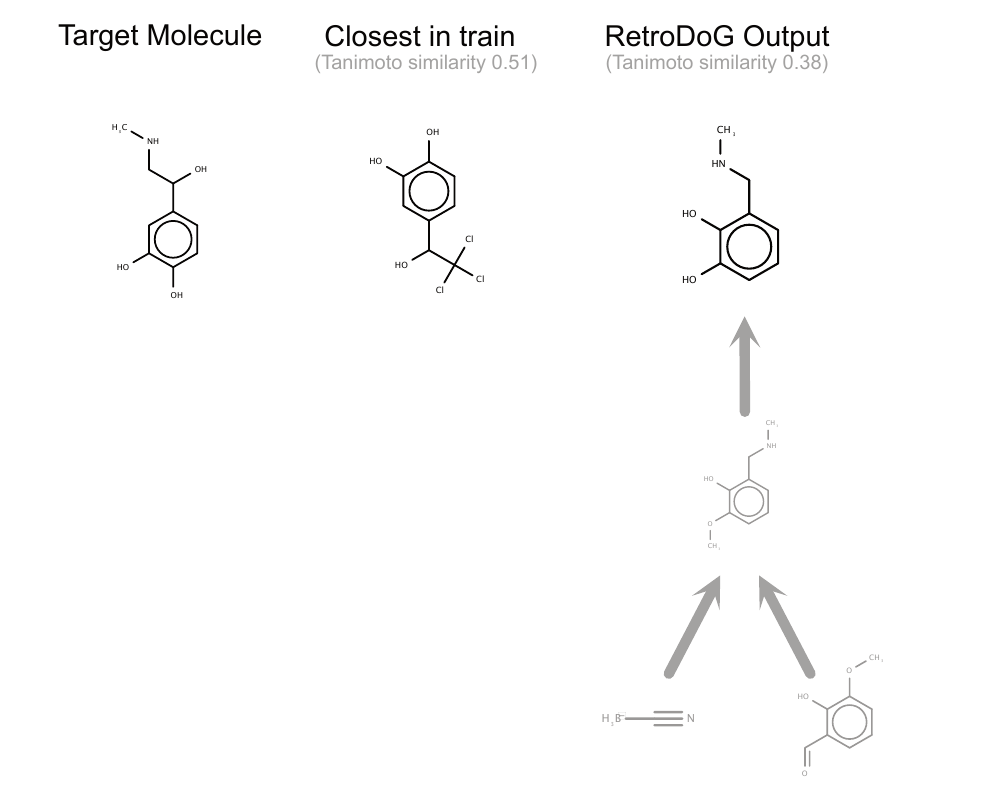}
\caption{
RetroDoG suggested DAG for epinephrine.
}
\label{fig:epinephrine}
\end{figure}

In order to qualitatively assess such a model we train RetroDoG on the same DAG dataset described
in Section \ref{sect:DatasetCreation}.  
At test time we sample 200 DAGs from our model and then sort them based on Tanimoto similarity (with Morgan fingerprints) between the final product molecule
and the original molecule
fed into RetroDoG, before picking the best one (note this uses only the input data and not the true target data).
In Figures \ref{fig:ribavirin}, \ref{fig:methotrexate} and \ref{fig:epinephrine} we show the result of this on three molecules taken from the WHO Essential Medicines \citep{who} list. 
Although we do not find a route to the exact molecule of interest, we often decode to similar final product molecules.
We find that RetroDoG finds molecules that are often as similar (measured again using the Tanimoto similarity between the fingerprints) as the best in our original training dataset.

We should point out that we do not expect this preliminary approach to currently be competitive with complex synthesis planning tools
such as the one proposed by \citet{segler2018planning}. These complex tools work in a top-down manner assessing 
many different routes. RetroDoG in contrast tries to construct the DAG in a bottom-up manner and is not able to roll back and 
adjust already made choices based on new data.
Having said that, we believe such an approach as RetroDoG may be worthy of future research interest,
and may for instance be useful in combination with more complex tools to amortize and reduce the cost of searching for synthetic routes.

\FloatBarrier

\section{Further Background}

This section contains some further background and context on molecule design, complimenting the introduction and related work sections in the main paper.

\label{sect:moreRelated}

\paragraph{Virtual screening and de novo design.} 
In computer-aided molecular discovery there are two main approaches to come up with the next molecules to make and test, 
virtual screening and de novo design \citep{hartenfeller2011enabling}.
In virtual screening (VS), large molecule libraries are pre-generated and then scored or searched on demand
, sometimes using ML models  \citep{muratov2020qsar}, requiring $\mathcal{O}(N)$ time and memory/storage.
However, given the combinatorial nature of molecules, which results in search spaces of the order $10^{24} - 10^{60}$, complete 
coverage of chemical space is neither feasible nor desirable \citep{walters1998virtual,shoichet2004virtual, pyzer2015high, van2019virtual,walters2018virtual,
reymond2012enumeration}.
Instead, the technique of \textit{de novo design} can be used, where a molecule construction algorithm is coupled with an optimization or
search procedure, and novel molecules are generated on the fly \citep{hartenfeller2011enabling, schneider2013novo}. 

When comparing different molecule generation algorithms, there is always a tradeoff between the coverage of chemical space,
and synthesizability, stability, and even to some extent the sensibleness of suggested structures. 
Synthesizability, stability and the sensibleness depend on the application area, are difficult to define, and can all appear subtle to non-chemists.
For example, although we can encode a simple notion of syntactic correctness through valency constraints 
(for instance having maximally four substituents at a carbon atom), these do not guarantee that a molecule is semantically valid.
Semantic validity is not trivial to define, and can be related to stability or the presence of undesirable substructures.
For instance, a molecule consisting of a chain of 10 nitrogen atoms with alternating single and double bonds would satisfy valency constraints,
but would unlikely be stable; it would decompose or explode immediately!

We can view VS as being on one end of this tradeoff between chemical coverage and synthesizability/stability. 
By using pre-defined virtual libraries, we are limited in chemical coverage but can ensure that all the molecules included are sensible and synthesizable.
This for instance is sometimes done by generating the library by iterating over a set of readily available starting materials, and combining them using virtual reaction schemes \citep{chevillard2015scubidoo}.

Inhabiting the other end of this tradeoff, would be a completely unconstrained de novo design algorithm,
allowing any connecting bond between any atoms to construct the molecular graph.
Although such an algorithm covers every conceivable molecule, it would generate structures not even meeting valency constraints.
This can be remedied by the hand-coding of expert rules, e.g. \citep{Luo1996-sr}.
However, arguably a more scalable approach was the development of algorithms that constructed molecules through combining together larger,
sensible fragments \citep{schneider2000novo} and later algorithms that constructed molecules iteratively by combining together
building blocks via virtual reaction schemes\citep{vinkers2003synopsis,hartenfeller2012dogs}.
However, even with these approaches there is still the problem of how best to search through this action space,
a difficult discrete optimization problem, and so these algorithms often resorted to optimizing greedily one-step at a time or using genetic algorithms.

\paragraph{ML design-and-search techniques for molecules}
A paradigm shift in molecule construction algorithms was achieved by the introduction of neural generative models,
in particular variational autoencoders on sequences \citep{Gomez-Bombarelli2018-ex}, and RNN-based language models
\cite{segler2018generating, olivecrona2017molecular}.
To perform optimization, autoencoder-based models map from a continuous latent space to the discrete molecule space,
allowing optimization to be performed in the continuous space, for instance using Bayesian optimization, leveraging local smoothness properties
and gradient information.
Autoregressive-style models can be used for optimization using reinforcement learning
\citep{segler2018generating,olivecrona2017molecular, neil_exploring_2018,you2018graph}. 
Some of these SMILES-based LSTMs have also been successfully used to prospectively suggest new molecules for lab-based experiments 
\citep{merk2018novo,merk2018tuning,yang2020discovery}.

Even though pre-training on large molecule datasets of known molecules mitigates the issue of generating unstable and unsynthesizable
molecules to a considerable extent, candidate molecules generated by current machine learning systems can still be difficult to synthesize
and sometimes contain unrealistic moieties \citep{hartenfeller2011enabling, brown2019guacamol, gao2020synthesizability}. 
This is particularly the case when optimizing.
So again the question becomes how can we build these constraints into our models?

\paragraph{Going further than molecular string based representations}
The work of \citet{Gomez-Bombarelli2018-ex} relied on string representations of molecular graphs (SMILES \citep{weininger1988smiles}), which
although flexible can lead to invalid molecules.
Extensions of this approach have therefore investigated representing molecules with grammars \citep{Kusner2017-ry, dai2018syntax},
and explicit constrained graph representations or molecular fragments \citep{Jin2018-jf,Jin2019-gm,Jin2018-aa,samanta2019nevae,Liu2018-ha, podda2020deep}.
However, although these particular extensions improve the validity of generated molecules, they do not explicitly consider synthesizability.

Generative ML models which construct molecules from building blocks via chemical reactions (e.g. \citep{bradshaw2019model, korovina2019chembo, horwood2020molecular, gottipati2020learning} and the work presented here)
can fix this problem by design. The constraint of requiring a
viable synthesis route offers a powerful form of regularization, discouraging  unstable structures.
Furthermore, it provides interpretability to chemists as well as a possible head start in making, and in
turn testing, any proposed molecules.

\renewcommand\refname{Appendix References}

\end{document}